\documentclass[10pt,twocolumn,letterpaper]{article}

\usepackage{cvpr}              

\usepackage{xcolor}
\usepackage{multirow}
\usepackage[percent]{overpic}
\usepackage{graphicx}
\usepackage{pifont}
\usepackage{float}
\usepackage{cuted}
\usepackage[symbol]{footmisc}

\newcommand{\greencheck}{\textcolor{green}{\ding{52}}}
\newcommand{\redx}{\textcolor{red}{\ding{56}}}

\definecolor{cvprblue}{rgb}{0.21,0.49,0.74}
\usepackage[pagebackref,breaklinks,colorlinks,allcolors=cvprblue]{hyperref}

\title{EasyHOI: Unleashing the Power of Large Models for Reconstructing Hand-Object Interactions in the Wild}

\def\paperID{12917} 
\def\confName{CVPR}
\def\confYear{2025}

\author{
Yumeng Liu$^{1,2}$
\quad
Xiaoxiao Long$^{3,*}$
\quad
Zemin Yang$^{2}$
\quad
Yuan Liu$^{3,4}$
\quad
Marc Habermann$^{5}$\\
\quad
Christian Theobalt$^{5}$
\quad
Yuexin Ma$^{2,*}$
\quad 
Wenping Wang$^{6}$\\
$^1$ The University of Hong Kong , $^2$ShanghaiTech University,\\
$^3$Hong Kong University of Science and Technology, 
$^4$Nanyang Technological University,\\
$^5$Max Planck Institute for Informatics, $^6$Texas A\&M University
}

\begin{document}
\maketitle
\begin{strip}
    \centering
    \vspace{-5.2em}
    \includegraphics[width=\linewidth]{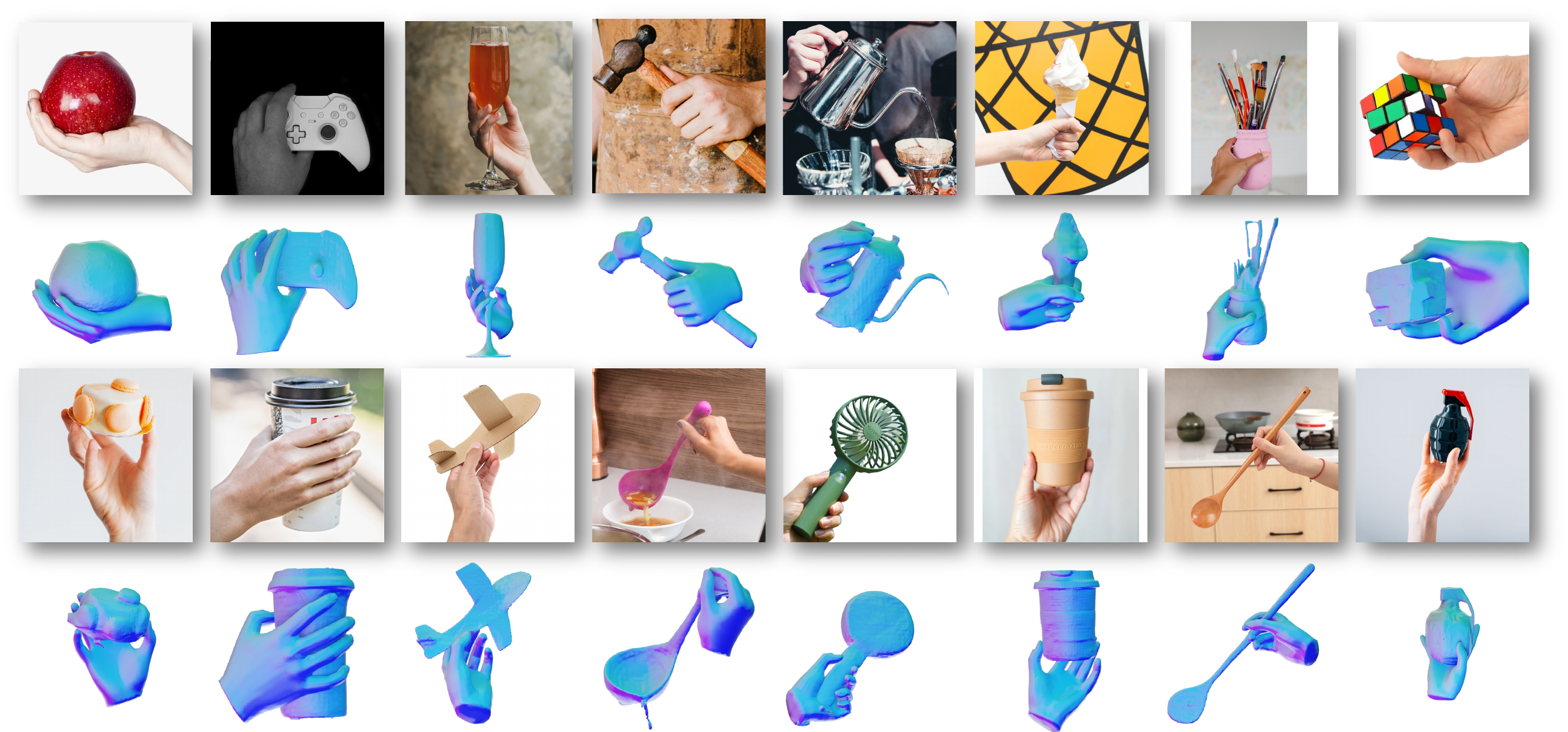}
    \vspace{-1em}
    \captionof{figure}{We leverage large models and prior-guided optimization to reconstruct physically plausible hand-object interactions from single-view in-the-wild images, effectively handling occlusions and diverse hand-object configurations.}
    \label{fig:teaser}
\end{strip}

\footnotetext{ * indicates the corresponding authors. }
\footnotetext{This work was finalized during the first author's visit to ShanghaiTech University.}
\begin{abstract}
Our work aims to reconstruct hand-object interactions from a single-view image, which is a fundamental but ill-posed task.
Unlike methods that reconstruct from videos, multi-view images, or predefined 3D templates, single-view reconstruction faces significant challenges due to inherent ambiguities and occlusions. These challenges are further amplified by the diverse nature of hand poses and the vast variety of object shapes and sizes.
Our key insight is that current foundational models for segmentation, inpainting, and 3D reconstruction robustly generalize to in-the-wild images, which could provide strong visual and geometric priors for reconstructing hand-object interactions.
Specifically, given a single image, we first design a novel pipeline to estimate the underlying hand pose and object shape using off-the-shelf large models. 
Furthermore, with the initial reconstruction, we employ a prior-guided optimization scheme, which optimizes hand pose to comply with 3D physical constraints and the 2D input image content.
We perform experiments across several datasets and show that our method consistently outperforms baselines and faithfully reconstructs a diverse set of hand-object interactions.
Here is the link of our project page: \url{https://lym29.github.io/EasyHOI-page/}.

\end{abstract}    
\section{Introduction}
Reconstructing \textbf{H}and-\textbf{O}bject \textbf{I}nteractions (HOI) from a single image is an essential task in computer vision and graphics, which plays important roles in many applications, such as human behavior understanding, Augmented Reality, teleoperation, and robotic grasping. 
This task aims to not only reconstruct the shapes of objects and hands but also to recover fine-grained hand-object interactions satisfying physical constraints. 

Single-view HOI reconstruction is quite challenging, since the monocular setting is naturally ill-posed and the interactions contain severe mutual and self-occlusions.
Many existing works~\cite{garcia2018first,tekin2019h+,hampali2020honnotate} typically simplify the HOI task by assuming the object is known.
However, such 3D template assumption is too strong and prevents these systems from generalizing to unknown objects. Some other works~\cite{chao2021dexycb,hampali2020honnotate,hasson19_obman} directly train neural networks for joint object and hand recovery on public datasets in an end-to-end manner.
However, since acquiring 3D annotations of HOI data is significantly difficult and expensive, the public datasets are low-quality and limited. The lack of high-quality HOI data proves to be the key challenge to boost the development of this task~\cite{cao2021reconstructing}. Overall, these methods unavoidably suffer from poor robustness and could not generalize to unseen scenarios.

Recent advancements in large models have shown promising potential in reconstructing both objects and hands. Notable studies in single-view 3D reconstruction models~\cite{liu2023syncdreamer,long2023wonder3d,hong2023lrm,li2023instant3d,xu2024instantmesh,tripo3d} have demonstrated remarkable capabilities in understanding complex geometries from limited visual input. Similarly, advances in hand estimation~\cite{pavlakos2024reconstructing,xu2022vitpose,simon2017hand} have shown significant progress in inferring intricate hand poses in complex scenarios. These parallel developments offer encouraging prospects for addressing the HOI task.  

A seemingly straightforward approach is independently reconstructing and merging hands and objects using pre-trained large models. However, this approach encounters three critical limitations: 1) \textbf{Inconsistent Coordinate Systems:} Object reconstruction methods generally rely on a canonical coordinate system, while hand reconstruction methods often adopt a camera-based system that aligns with the input viewpoint. This incompatibility leads to misaligned results when merged directly.
2) \textbf{Estimation Inaccuracies:} The reconstruction methods for, both, objects and hands frequently exhibit significant inaccuracies, often resulting in spatial misalignments that fail to reflect the actual observed interactions.
3) \textbf{Impact of Occlusions:} Inferring geometry and interactions for occluded regions from a single view poses significant challenges. The inability to effectively reconstruct these unobserved areas leads to physically implausible results and unrealistic interaction estimates.

To achieve robust and high-quality Hand-Object Interaction (HOI) estimation, we propose leveraging the strong priors of multiple large models to reconstruct both the object and hand. Additionally, we introduce a novel prior-guided optimization framework to jointly optimize these reconstructions, ensuring physically plausible HOI results.
Our proposed optimization framework comprises three key stages:
1) \textbf{Camera System Setup:} Using differentiable rendering techniques, we align the reconstructed object and hand within a unified coordinate system, guided by segmentation priors to ensure coherent positioning.
2) \textbf{HOI Contact Alignment:} We identify potential 3D interaction regions between the object and hand, applying the Iterative Closest Point (ICP) algorithm to register their shapes. This step yields an initial, approximate alignment that captures the interaction.
3) \textbf{Hand Parameter Refinement:} To enhance physical plausibility, we refine hand parameters using a combination of loss functions, including segmentation mask loss, penetration loss, contact loss, and regularization loss. This multi-loss approach optimizes the hand's interaction with the object for a realistic and accurate result with limited visual observation.

Compared to the prior works, our system presents strong generalization and robustness on diverse in-the-wild images and could produce plausible HOI reconstructions with only single image as input.
Extensive experiments have been conducted to validate the strong generalization and robustness of our method.

\section{Related Works}
\subsection{Hand-Object Interactions Reconstruction}
Reconstructing hand-and-object interactions is a challenging problem that has attracted significant attention in recent years. Many reconstruction methods rely on richer input sources, such as videos~\cite{hasson20_handobjectconsist,huang2022hhor,ye2023vhoi,fan2024hold} and object templates~\cite{garcia2018first,hamer2010object,hampali2020honnotate,romero2010hands,sridhar2016real,tekin2019h+,tzionas2016capturing,tzionas20153d,chen2021joint,gkioxari2018detecting,liu2021semi,Shan20,tekin2019h+,brahmbhatt2020contactpose,cao2021reconstructing,corona2020ganhand,grady2021contactopt,hamer2010object,pham2017hand,tzionas2016capturing, zhang2020perceiving}, restricting their use in real-world applications.
To achieve model-free reconstruction from single view, some methods deform a sphere mesh~\cite{hasson19_obman} or learn an implicit shape field for objects~\cite{karunratanakul2020grasping,ye2022hand, zhang2024moho, chen2022alignsdf, chen2023gsdf}. However, these approaches are often trained on limited datasets, leading to reduced effectiveness on objects that differ from the training data. In contrast, our method leverages advanced large-scale models to significantly enhance generalization capabilities.

\subsection{Single-view 3D Reconstruction}
Single-view 3D reconstruction has been a long-standing focus in computer vision and graphics, serving as a foundational step for numerous applications. Early approaches~\citep{tatarchenko2019single,fu2021single,kato2019learning,li2020self,fahim2021single,niemeyer2021giraffe,chan2022efficient,gu2021stylenerf,schwarz2020graf} applied neural networks to reconstruct 3D shapes from single-view images through regression~\citep{li2020self}, retrieval~\citep{tatarchenko2019single}, or NeRF-based GANs~\cite{niemeyer2021giraffe}. Recent trends leverage diffusion models using strategies such as Score-Distillation Sampling~\cite{poole2022dreamfusion,liu2023zero123,liang2023luciddreamer} and multiview diffusion~\cite{liu2023syncdreamer,long2023wonder3d,liu2023zero123,shi2023MVD}. 
Additionally, recent Large Reconstruction Models (LRMs)~\cite{hong2023lrm,wang2023pflrm,wang2024crm,tang2024lgm} employ transformers to reconstruct object meshes directly from input images. Our method incorporates an LRM model as well but focuses on single-view reconstruction within the context of human-object interactions (HOI).
\begin{figure*}[h]
    \centering
    \includegraphics[width=\textwidth]{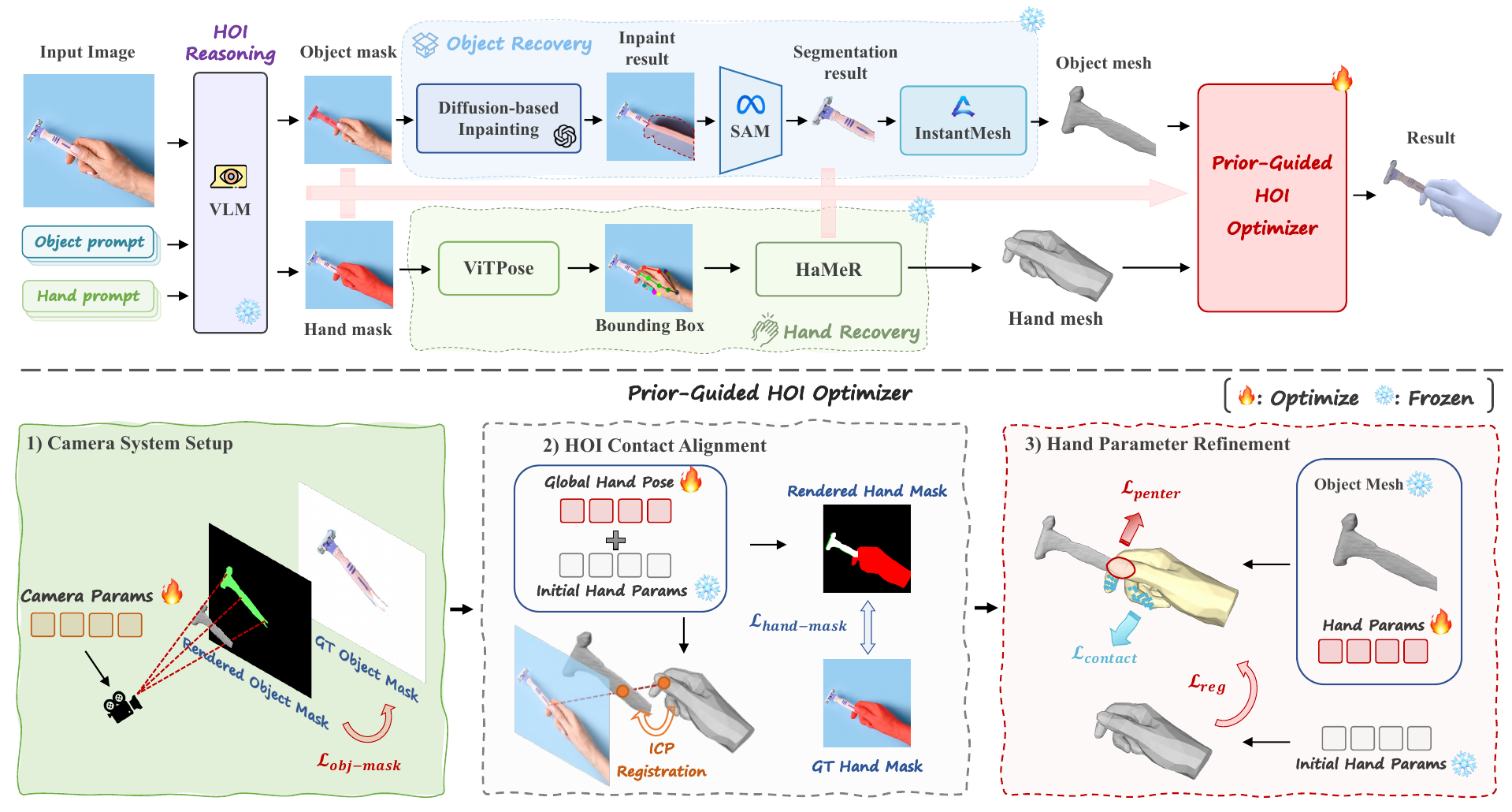}
    \vspace{-2em}
    \caption{
    The illustration of our pipeline. We first extract hand and object masks through HOI reasoning for object reconstruction and recovering hand mesh from the input image. With these initial reconstructions, we employ a three-stage prior-guided optimizer to establish a camera system for object, align hand and object by analyzing contact points, and finally refines hand parameters to ensure physical plausibility.
    }
    \vspace{-1.5em}
    \label{fig:pipeline}
\end{figure*}

\subsection{Image-based Hand Pose Estimation}
Image-based hand pose estimation aims to recover hand pose, including keypoints, parametric shape parameters, or geometries, from input images. Early methods~\cite{iqbal2018hand,mueller2018ganerated,mueller2019real,panteleris2018using} primarily focused on estimating 2D or 3D hand keypoints. Recent single-view hand pose estimation studies have benefited from advancements in human mesh recovery, with the MANO parametric hand model~\cite{romero2022embodied} emerging as a powerful tool for extracting hand pose and shape parameters from RGB images~\cite{baek2019pushing,boukhayma20193d,zhang2019end,sridhar2015fast,zhou2020monocular}. These methods either directly regress MANO parameters~\cite{boukhayma20193d,baek2019pushing} or optimize shape fitting to the images~\cite{zhang2019end,sridhar2015fast,zhou2020monocular}.
Our approach also involves hand pose estimation, but we address it within the context of human-object interaction, where reconstructed objects introduce additional constraints to refine the hand meshes.
\section{Methodology}
\subsection{Problem Formulation}
Given an input image $I$ depicting hand-object interaction, our goal is to reconstruct the 3D object shape and determine its relative pose with respect to the hand. This task requires optimizing several interrelated components.
First, we obtain initial reconstructions of both the object mesh $\Omega_o$ and hand mesh $\Omega_h$. For hand representation, we employ the MANO parametric model~\cite{romero2022embodied}, which is characterized by two key components: the global 6D pose $\phi_h \in \mathbb{R}^6$ and the articulated pose parameters $\theta_h \in \mathbb{R}^{45}$.
The subsequent optimization involves estimating camera parameters, including intrinsic matrix $K$ and extrinsic parameters $[R | t]$ for object-image alignment, and refining hand parameters $(\phi_h, \theta_h)$.

\subsection{Pipeline Overview}
As shown in~\cref{fig:pipeline}, our pipeline mainly consists of two stages: 
1) \textbf{Initial Reconstruction of Hand and Object.} We first employ LISA~\cite{lai2023lisa} to segment hand and object masks. A diffusion model~\cite{ye2023affordance} then removes and inpaints the hand region, followed by SAM~\cite{kirillov2023segment} segmenting the complete object. Finally, InstantMesh~\cite{xu2024instantmesh} reconstructs the object mesh while HaMeR~\cite{pavlakos2024reconstructing} simultaneously reconstructs the hand mesh.
2) \textbf{Hand-Object Interaction Optimization.} Since the initial hand and object reconstructions are obtained separately, they may be inconsistent with the input image. 
To overcome this challenge, we formulate a prior-guided optimization framework for HOI tasks that progressively refines the reconstruction in a coarse-to-fine manner.

\begin{figure}[htp]
    \centering
    \begin{overpic}[width=\linewidth]{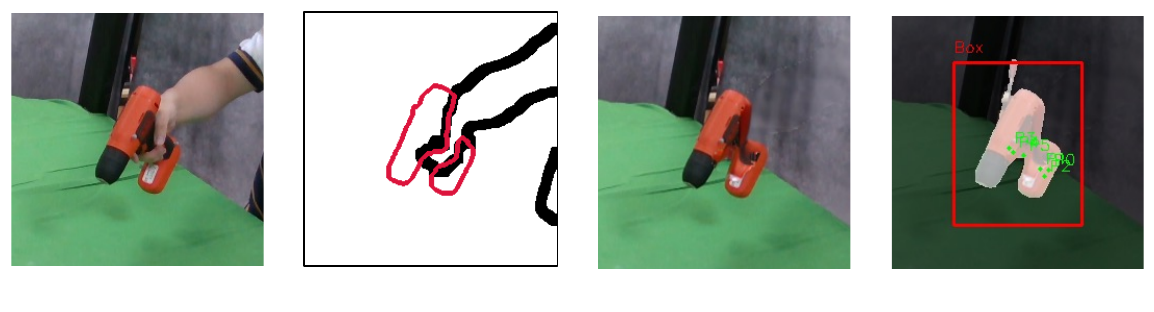}
        \put(7,0){\scriptsize (a) Input}
        \put(30,0){\scriptsize (b) Contours}
        \put(56,0){\scriptsize (c) Inpaint}
        \put(78,0){\scriptsize (d) Segmentation}
    \end{overpic}
    \vspace{-1em}
    \caption{ Illustration of the segmentation process after inpainting. (a) Original input image. (b) Hand and object contours, showing the object split into two disconnected regions due to hand occlusion. (c) Inpainted image with the hand region removed. (d) Object segmentation results, consisting of sampled points and a bounding box, used as prompts to segment the inpainted image.}
    \vspace{-1.5em}
    \label{fig:seg_inpaint}
\end{figure}

\subsection{Initial Reconstruction of Hand and Object}
\noindent \textbf{Hand-Object Interaction Reasoning.}
\label{subsec:hoi_reason}
We first segment the regions of interest - the hand and its interacting object - from the input image, which serves as the foundation for subsequent reconstruction. 
To handle images that may contain multiple objects, we employ LISA~\cite{lai2023lisa}, a vision-language model for segmentation task, to obtain semantic segmentation masks for the relevant hand-object interaction.

While LISA effectively segments hands and objects in most cases, it sometimes produces redundant masks for visually similar objects that are not actually being interacted with. To address this, we propose a contour-guided filtering strategy that discards objects whose contours are not neighboring the hand's. Finally we obtain accurate segmentation masks for the interacting objects, denoted as $M_o$. 
Additional details are presented in the supplementary material. 
\noindent \textbf{Hand Reconstruction.} 
We employ a recent transformer-based approach, HaMeR~\cite{pavlakos2024reconstructing} to recover the hand pose from the input image. 
First, ViTPose~\cite{xu2022vitpose} is utilized to estimate the 2D keypoints of the hands and compute the bounding box based on these keypoints, which helps crop the regions of interest (ROI) from the whole input image. 
Afterward, the hand parameters $(\phi_h, \theta_h)$ are predicted from the bounding box via HaMeR. 

\noindent \textbf{Object Reconstruction.}
Due to the occlusion caused by hand, the interacted object is often visually incomplete in the input image and therefore could only produce distorted 3D geometry of the object. 
To address this problem, as the example shown in \cref{fig:seg_inpaint}, our reconstruction pipeline first employs a diffusion model~\cite{ye2023affordance, nichol2021glide} to recover the complete object appearance by removing occlusions. 
As a by-product, we could obtain the segmentation mask of the inpainted object with complete appearance, denoted as $\hat{M}_o$.

Finally we apply a large reconstruction model, InstantMesh~\cite{xu2024instantmesh}, to reconstruct the object's full geometry and conduct a post-processing to make the generated mesh watertight. A detailed explanation can be found in the supplementary material.

\subsection{Hand-Object Interaction Optimization}
Although initial reconstructions of the hand and object are obtained, their separate processing and differing coordinate systems result in unrealistic spatial relationships, such as excessive distance or interpenetration. Furthermore, depth and scale ambiguities in monocular images exacerbate these issues. To resolve this, we propose a three-step optimization scheme to reconstruct a realistic interaction.

\noindent\textbf{1) Camera System Setup.}
The problem of aligning the two systems of hand and object can be simplified by using the reconstructed object's coordinate frame as the global reference, and then estimating the camera parameters of the input image defined in the object coordinate system. 
Using a differentiable renderer $\Psi$, we would like to obtain the optimal camera parameters ($K$, $[R | t]$) by minimizing the soft IoU loss between rendered object mask $M_o^{r} = \Psi(\Omega_o,K,R,t)$ and ground truth mask $\hat{M}_o$ (the segmentation mask of the inpainted object with complete appearance), as shown in~\cref{eq:object_iou_loss}.
\begin{equation}
    \label{eq:object_iou_loss}
    \mathcal{L}_{\text{obj-mask}} = \operatorname{IOU}(M_o^{r},\hat{M}_o) = \frac{M_o^{r} \cdot \hat{M}_o}{|M_o^{r}|+|\hat{M}_o|- M_o^{r} \cdot \hat{M}_o},
\end{equation}
where $|.|$ denotes the sum of all elements in the mask.

However when initial and ground truth masks don't overlap, the IoU loss becomes ineffective since its gradient is zero. We address this by incorporating Sinkhorn distance~\cite{cuturi2013sinkhorn}, a regularized Wasserstein distance that measures the minimal cost of transforming the rendered mask distribution to match the ground truth. Let $M_{\alpha}$ and $M_{\beta}$ represent two normalized mask images, treated as discrete distributions supported on a finite grid. The Wasserstein distance between them is defined by:
\begin{equation}
\begin{split}
&W(M_{\alpha}, M_{\beta}) = \\
&\left( \inf_{\gamma \in \Pi(M_{\alpha}, M_{\beta})} \sum_{i, j} \sum_{k, l} ||(i,j) - (k,l)||^2 \gamma_{ijkl} \right)^{1/2},
\end{split}
\end{equation}
where $\gamma_{ijkl}$ represents the transport flow from position $(i, j)$ in $M_{\alpha}$ to position $(k, l)$ in $M_{\beta}$, and $\Pi(M_{\alpha}, M_{\beta})$ denotes the set of all possible transport flows. We employ the Sinkhorn-Knopp algorithm~\cite{cuturi2013sinkhorn} to compute the optimal transport loss:
\begin{equation}
\mathcal{L}_{\text{OT}} = W (\dfrac{M_o}{|M_o|}, \dfrac{\hat{M_o}}{|\hat{M_o}|}).
\end{equation}
Combining this with IoU loss yields optimal camera parameters $\bar{K}$ and $[\bar{R}|\bar{t}]$, ensuring robust alignment even for non-overlapping masks.

\noindent\textbf{2) HOI Contact Alignment.}
Since we adopt the object’s coordinate system as the global reference, the object is positioned at the system origin, allowing us to focus solely on optimizing the hand parameters for hand-object interaction (HOI) reconstruction.
In this stage, we retain the articulated hand pose $\theta_h$ as estimated by HaMeR, and optimize the hand’s global 6D pose $\phi_h$ to bring the hand and object into approximate contact. The optimization alternates between two steps: \textbf{\textit{Mask-Constrained Optimization}} and \textbf{\textit{Contact-Based Registration}}, iteratively improving contact alignment.

\noindent\textbf{\textit{(a) Mask-constrained Hand Pose Optimization.}}
We first need an initial 2D alignment between the estimated hand configuration and the input image, leveraging hand segmentation as supervision. The differentiable renderer \(\Psi\) generates a hand mask \(M_h^{r} = \Psi(\theta_h, \phi_h, \bar{K}, \bar{R}, \bar{t})\). We then compute the soft IoU loss between this rendered mask \(M_h^{r}\) and the ground truth hand mask \(M_h\) obtained from input image segmentation.
\begin{equation}
    \label{eq:hand_iou_loss}
    \mathcal{L}_{\text{hand-mask}} = \operatorname{IOU}(M_h^{r}, M_h) =  \frac{M_h^{r} \cdot {M}_h}{|M_h^{r}|+|{M}_h|- M_h^{r} \cdot {M}_h}.
\end{equation}
The hand mask loss ensures hand alignment with the input image in the 2D plane. However, differentiable rendering only provides gradients parallel to the image plane, not in the camera viewing direction. This fundamental limitation makes it challenging to achieve precise 3D alignment between the hand and the object.

\begin{figure}[tp]
    \centering
    \includegraphics[width=0.9\linewidth]{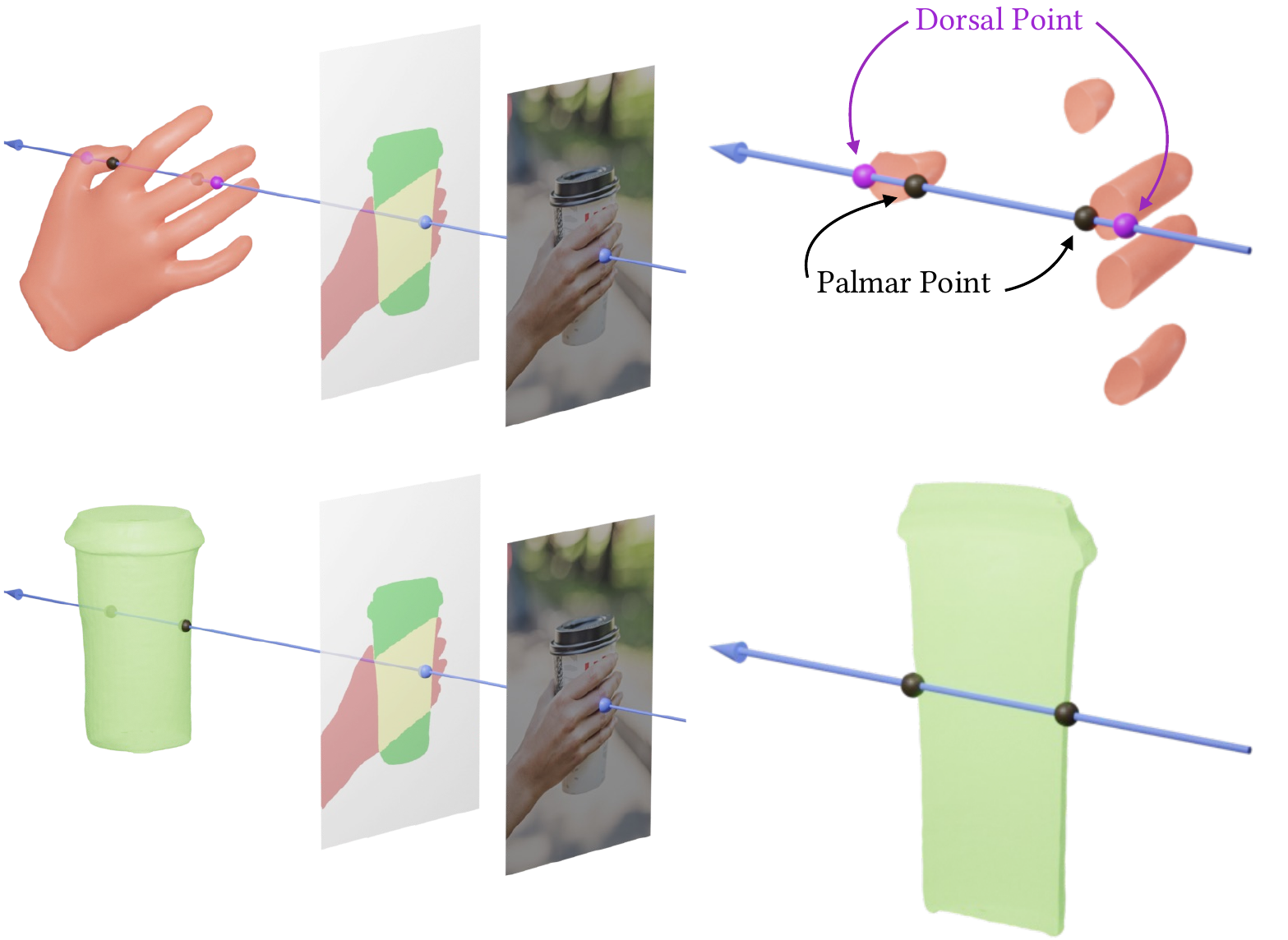}
    \vspace{-1em}
    \caption{Converting 2D contact regions to 3D contact points. Rays emitted from contact mask pixels intersect object and hand geometries. Contact point candidates are constrained to the extremal ray intersections: nearest or farthest points relative to the camera for the object, and palmar-side extremal points for the hand.}
    \label{fig:contact-align}
    \vspace{-1.5em}
\end{figure}

\noindent\textbf{\textit{(b) ICP-based Hand-Object Registration.}} To achieve accurate hand-object alignment, analyzing their contact relationships is essential. Mutual occlusions provide inherent contact cues, which we use to detect 2D contact regions (detailed methodology in the supplementary material).  These 2D contact regions are then converted to 3D contact points through ray-casting. As shown in Fig.~\ref{fig:contact-align}, ray-casting from image pixels creates multiple intersection points due to monocular ambiguity. For objects, we select the nearest and farthest intersection points along each ray. For hands, we focus only on the palm side as the functional grasping area, excluding the dorsal side (the back of the hand) using pre-marked regions on the MANO template. We then identify valid hand contact points by first keeping only palmar intersections, then selecting the nearest and farthest points along each ray.

Once all potential contact points on both the hand and the object are identified, we apply the Iterative Closest Point (ICP) method to compute the optimal hand translation, aligning the contact points and providing a rough estimation of the hand's pose.

\begin{figure*}
\centering
    \includegraphics[width=0.9\textwidth]{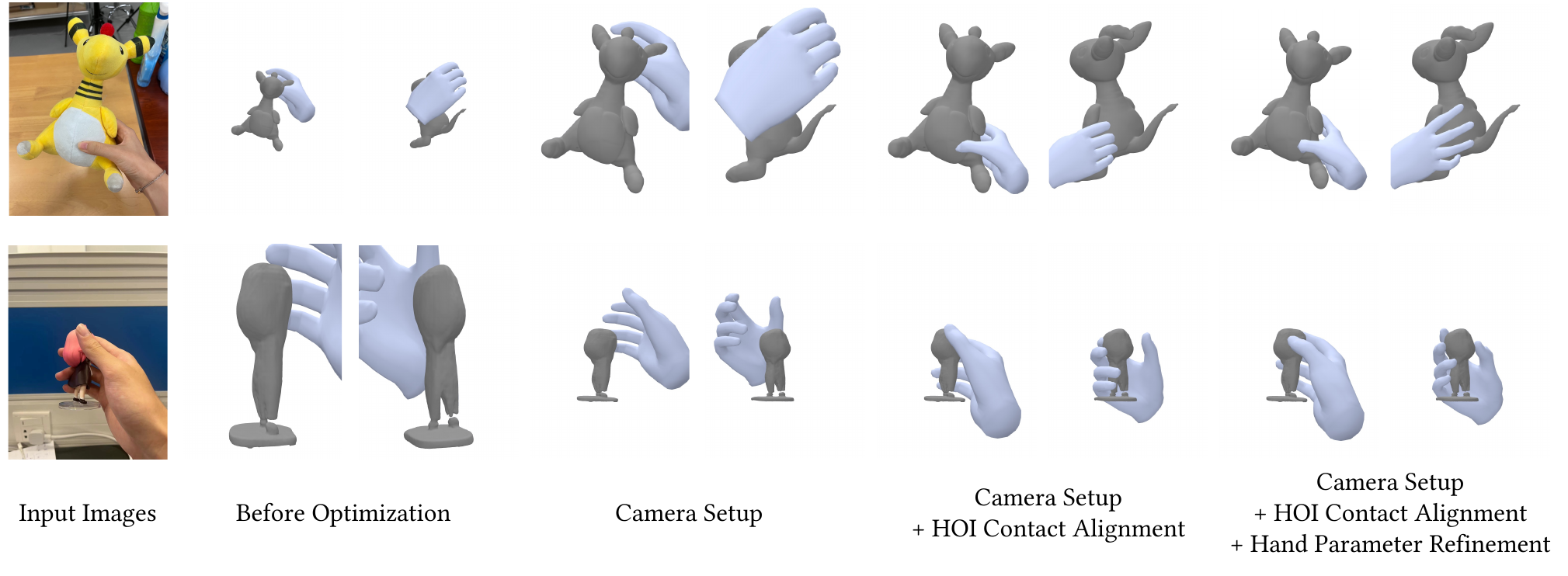}
    \vspace{-1.5em}
    \caption{We perform ablation studies by examining the outputs at each processing stage. Starting with reconstruction results from foundational models, we present the progressive improvements through Camera Setup, HOI Contact Alignment, and Hand Parameter Refinement.}
    \label{fig:hand_optim_stage}
    \vspace{-1.5em}
\end{figure*}

\noindent\textbf{3) Hand Parameter Refinement.}
While initial global pose optimization provides coarse alignment, finger configurations remain unadapted to the object's geometry. We then jointly optimize the hand's global pose $\theta_h$ (6 DoF) and articulation parameters $\phi_h$ (45 DoF) to achieve physically plausible, penetration-free configurations. The objective function is defined as:
\begin{equation}
    \mathcal{L}_{\text{hand}} = \lambda_1 \mathcal{L}_{\text{hand-mask}} + \lambda_2 \mathcal{L}_{\text{penetr}}  + \lambda_3 \mathcal{L}_{\text{contact}} + \lambda_4 \mathcal{L}_{\text{reg}},
\label{eq:total_loss}
\end{equation}
where \(\mathcal{L}_{\text{hand-mask}}\) represents the IoU loss of the hand mask, as shown in~\cref{eq:hand_iou_loss}, \(\mathcal{L}_{\text{penetr}}\) is the penetration loss to prevent hand-object penetration, \(\mathcal{L}_{\text{contact}}\) encourages a reasonable hand-object contact relationship, and \(\mathcal{L}_{\text{reg}}\) is a regularization loss that ensures the articulated hand pose remains close to the result from HaMeR. The parameters \(\lambda_1, \lambda_2, \lambda_3\) and \(\lambda_4\) balance the different losses.

\noindent
\textbf{\textit{Penetration Loss.}} We introduce a penetration loss in this stage, defined as the mean distance from hand vertices inside the object to its boundary, to discourage intersections. 
\begin{equation}
    \label{eq:penetration_loss}
    \mathcal{L}_{\text{penetr}} = \frac{1}{N} \sum_{v\in\mathcal{H}} \max(0, -d(v)),
\end{equation}
where $d(\cdot)$ is the signed distance function defined around the object mesh, $\mathcal{H}$ represents the hand mesh and $v$ is arbitrary vertex of the hand mesh.

\noindent
\textbf{\textit{Contact Loss.}}
To ensure sufficient contact and stability of grasps, we incorporated a contact loss. We use the contact zones $\mathcal{H}_C$ introduced by ObMan~\cite{hasson19_obman}, which are the regions of hand mesh that frequently make contact during grasping. We then calculate the total distance from the exterior points sampled in the regions to the object surfaces.
\begin{equation}
    \label{ch2:eq:contact_loss}
    \mathcal{L}_{\text{contact}} = \sum_{v\in\mathcal{H}_C} \max(0, d(v)).
\end{equation}

\noindent
\textbf{\textit{Regularization Loss.}}
To ensure that the optimized hand pose does not deviate significantly from the initial estimates, we employ a regularization loss. Specifically, we utilize an L1 loss to compare the optimized pose with the initial pose $\theta_h^0$ estimated by the HaMeR system. 
\begin{equation}
    \label{ch2:eq:regularization_loss}
    \mathcal{L}_{\text{reg}} = \|\theta_h - \theta_h^0\|_1.
\end{equation}
This regularization maintains a balance between accuracy and realism, preventing excessive deviations from the HaMeR predictions in the pose adjustments. 

\section{Experiments}
\begin{table*}[htpb]
    \centering
    \resizebox{\linewidth}{!}{
\begin{tabular}{cccccc|ccccc|ccccc}
\hline 
& \multicolumn{5}{c}{ Arctic } & \multicolumn{5}{c}{ OakInk } & \multicolumn{5}{c}{ DexYCB } \\
\hline & F5 $\uparrow$ & $\mathrm{F} 10 \uparrow$ & $C.D.\downarrow$ & S.D.$\downarrow$ & I.V.$\downarrow$
& F5 $\uparrow$ & $\mathrm{F} 10 \uparrow$ & $C.D.\downarrow$ & S.D.$\downarrow$ & I.V.$\downarrow$
& F5 $\uparrow$ & $\mathrm{F} 10 \uparrow$ & $C.D.\downarrow$ & S.D.$\downarrow$ & I.V.$\downarrow$\\
\hline 
IHOI~\cite{ye2022hand} & 0.083 & 0.164 & 1.375 &2.79 &4.94 & 0.097 & 0.152 & 1.742 &3.14 &4.66 &0.084 &0.143 & 1.897 & 2.59 &4.93\\
AlignSDF~\cite{chen2022alignsdf} &0.102  &0.196  &1.289 &2.46 &4.73 &0.095  &0.148  &1.813 &3.27 &4.61  &- &- &- &- &-\\
gSDF~\cite{chen2023gsdf} &0.115  &0.247  &1.247 &2.31 &4.89 &0.106  &0.173  &1.992 &3.15 &4.16 &- &- &- &- &-\\
MOHO~\cite{zhang2024moho} &0.072  &0.136  &12.878 &3.94 &4.72 &0.175  &0.323  &3.883 &3.55 &4.86  &0.119 &0.249 &1.695 &2.62 &4.61 \\
Ours & \textbf{0.155} & \textbf{0.272} & \textbf{1.089} &\textbf{2.25} &\textbf{4.67} & \textbf{0.247} & \textbf{0.445} & \textbf{1.035} & \textbf{3.08} & \textbf{4.11} &\textbf{0.134} &\textbf{0.253} & \textbf{1.628} &\textbf{2.43} & \textbf{4.52}\\
\hline
\end{tabular}
    }
\vspace{-0.8em}
\caption{Quantitative evaluation for object quality in HOI reconstruction. Since AlignSDF and gSDF were trained on DexYCB, we exclude their DexYCB results from our zero-shot comparisons. The metrics F5 and F10 measure the F score of points from reconstructed object within 5mm and 10mm of the GT object, respectively. The metric $C.D.$ denotes the Chamfer Distance between reconstructed object and GT object, $S.D.$ denotes Simulation Displacement(in $cm$) and $I.V.$ represents Intersection Volume(in $cm^3$).}
\label{tab:object_recon}
\vspace{-1em}
\end{table*}

\begin{table}[htpb]
    \centering
\resizebox{0.9\linewidth}{!}{
    \begin{tabular}{ccc|cc}
    \hline 
    & \multicolumn{2}{c}{ Arctic } & \multicolumn{2}{c}{ OakInk }  \\
    \hline 
    & MPVPE$\downarrow$ & MPJPE$\downarrow$ & 
 MPVPE$\downarrow$ & MPJPE$\downarrow$ \\
    \hline 
    IHOI &1.36 &1.57 &1.36  &1.32  \\
    AlignSDF &1.38 &1.36  &1.25  &1.26  \\
    gSDF &1.34 &1.37 &1.27  &1.14  \\
    MOHO &1.38 &1.35 &1.27  &1.33  \\
    HaMeR &1.14 &1.05 &1.06  &1.03  \\
    Ours (Pred-object) & 1.48 &1.36 & 1.19 & 1.22 \\
    Ours (GT-object) &\textbf{1.12} &\textbf{0.95} &\textbf{0.86} & \textbf{0.77}\\
    \hline
    \end{tabular}
}
\vspace{-0.5em}
\caption{Hand pose accuracy in HOI reconstruction is evaluated using MPVPE (Mean Per Vertex Position Error) and MPJPE (Mean Per Joint Position Error), both measured in centimeters (cm).}
\vspace{-1.5em}
    \label{tab:hand_metric}
\end{table}

\subsection{Experimental Setup}
Our method is implemented on an Ubuntu server equipped with an NVIDIA A40 GPU. For the camera system setup, we use a learning rate of $10^{-2}$ for optimizing camera parameters, terminating when $\mathcal{L}_{\text{obj-mask}} < 0.1$ or after 1000 iterations. In human-object interaction (HOI) contact alignment, global hand parameters are optimized with a learning rate of $10^{-2}$. Hand-object registration are performed only at the 100th and 200th iterations, mask-constrained hand pose optimization 
continue until $\mathcal{L}_{\text{hand-mask}} < 0.1$ or 1000 iterations. During hand parameter refinement, we set learning rates of $10^{-4}$ for the global hand pose and $10^{-2}$ for articulated pose. The loss weights in~\cref{eq:total_loss} are set to $\lambda_1 = 5$, $\lambda_2=10$, $\lambda_3=5$, and $\lambda_4=0.1$.

\subsection{Datasets and Baselines} 
To evaluate our method’s generalization capability, we conduct a zero-shot comparison with IHOI~\cite{ye2022hand}, AlignSDF~\cite{chen2022alignsdf}, gSDF~\cite{chen2023gsdf}, and MOHO~\cite{zhang2024moho}. Specifically, we test on three public datasets: Arctic~\cite{fan2023arctic}, OakInk~\cite{YangCVPR2022OakInk}, and DexYCB~\cite{chao:cvpr2021}. These datasets contain videos with 3D hand-object pose and shape annotations: Arctic includes 11 articulated objects, OakInk features 100 diverse objects, and DexYCB presents 20 distinct YCB-video objects. To ensure valid grasping in the images, we use a force closure tester~\cite{liu2021synthesizing} to filter out instances where the hand and object are not in contact. We then randomly select 500 images from each dataset for evaluation.
All datasets represent unseen domains for IHOI and MOHO. For AlignSDF and gSDF, DexYCB is part of their training data, so we exclude it from testing these methods to ensure a fair comparison.

\begin{figure*}[htp]
    \centering
    \includegraphics[width=0.9\linewidth]{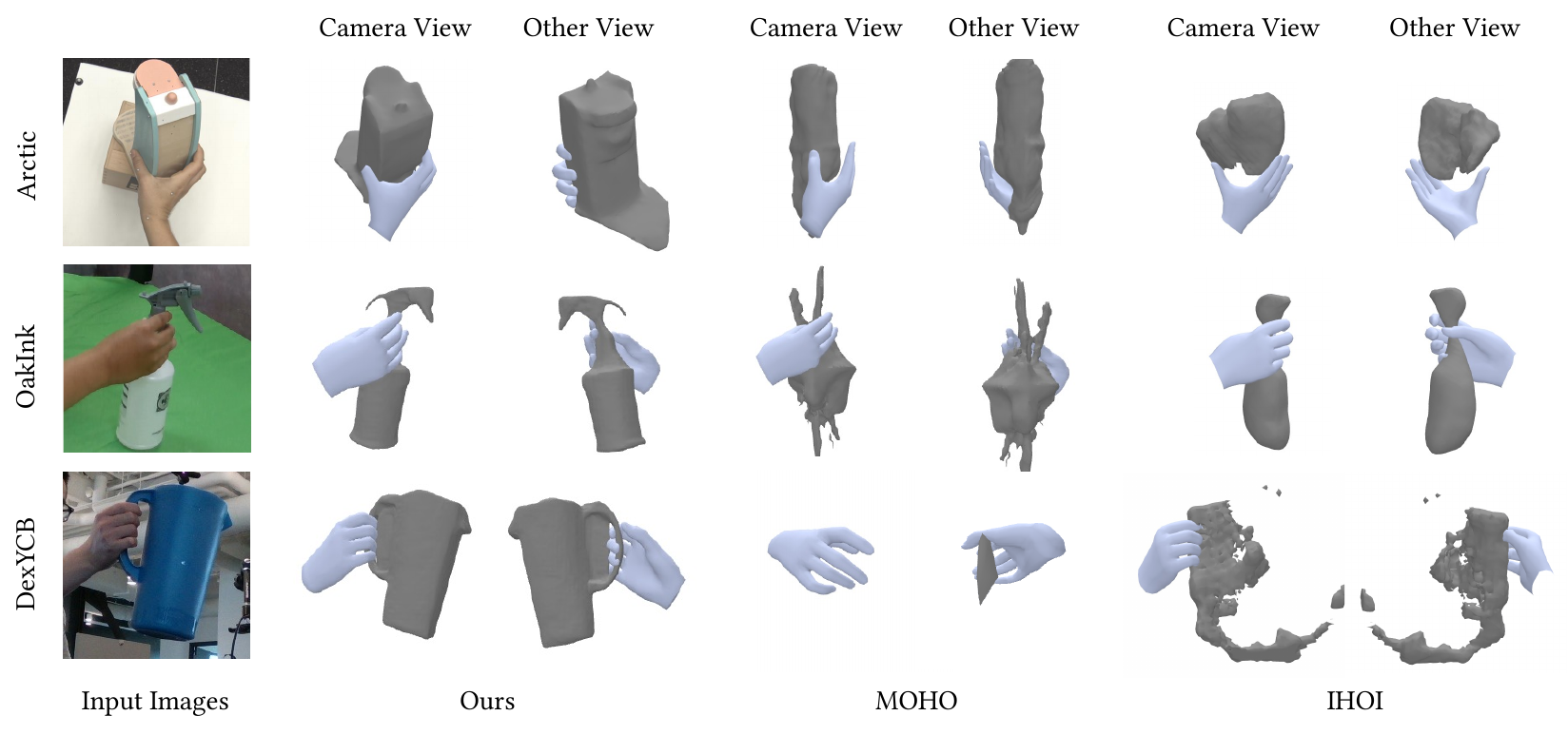}
    \vspace{-1.5em}
    \caption{This gallery showcases the outcomes of our hand-object reconstruction on three public dataset Arctic, OakInk and DexYCB (more results in the supp.). The first column is the input image, we present the camera view and another view to display the reconstructed HOI meshes. }
    \label{fig:gallery_3datasets}
    \vspace{-1.5em}
\end{figure*}

\begin{table*}[htp]
    \centering
    \resizebox{\linewidth}{!}{
    \begin{tabular}{ccccccc|cccc|cccc}
        \hline 
        & & & \multicolumn{4}{c}{ Arctic } & \multicolumn{4}{c}{ OakInk } & \multicolumn{4}{c}{ DexYCB } \\
        \hline
        Camera & HOI Contact  & Hand Parameter & \multicolumn{2}{c}{ 
        S.D.$\downarrow$
        } & \multicolumn{2}{c|}{ I.V.$\downarrow$} & \multicolumn{2}{c}{  S.D.$\downarrow$} & \multicolumn{2}{c|}{ I.V.$\downarrow$} & \multicolumn{2}{c}{S.D.$\downarrow$} & \multicolumn{2}{c}{I.V.$\downarrow$}
        \\
        Setup & Alginment & Refinement & Mean & Std & Mean & Std & Mean & Std & Mean & Std& Mean & Std & Mean & Std \\
        \hline 
        \redx & \redx & \redx & 5.79  & 6.35  & \textbf{0.49} & \textbf{1.63} & 5.81 & 5.94 & \textbf{0.47} & \textbf{1.82} & 6.32 &6.29 &\textbf{0.44} &\textbf{1.76} \\
        \greencheck & \redx & \redx & 5.79  & 6.35  & 0.49 & 1.63 & 5.81 & 5.94 & 0.47 & 1.82 & 6.32 &6.29 &0.44 &1.76 \\
        \greencheck & \greencheck & \redx & 2.87  & 2.84  & 8.36 & 9.06 & 3.57 & 3.66 & 10.13 & 11.74 & 2.93 &2.86 &9.64 &9.75 \\
        \greencheck & \greencheck & \greencheck & \textbf{2.25}  & \textbf{2.09}  & 4.67 & 4.54 & \textbf{3.08} & \textbf{2.92} & 4.11 & 4.77 & \textbf{2.43} & \textbf{2.41} & 4.52 & 4.64\\
        
        \hline
    \end{tabular}
    }
    \caption{Ablation study for the HOI prior-guided optimization scheme.}
    \label{tab:ablation-optim-comp}
\end{table*}

\subsection{Evaluation Criteria}
To evaluate object reconstruction quality, we use the \textbf{Chamfer Distance} to measure the discrepancy between predicted outputs and ground truth. To minimize the impact of outliers, we also report the \textbf{F-score} at 5mm and 10mm thresholds. For assessing hand-object interaction quality, we calculate the \textbf{Intersection Volume} between the hand and object models. Additionally, we perform a simulation with a fixed hand pose and gravity applied to the environment. By measuring the distance the object falls over 1,000 simulation steps, termed \textbf{Simulation Displacement}, we effectively evaluate grasp stability.

\begin{figure*}[t]
    \centering
    \begin{overpic}[width=\textwidth]{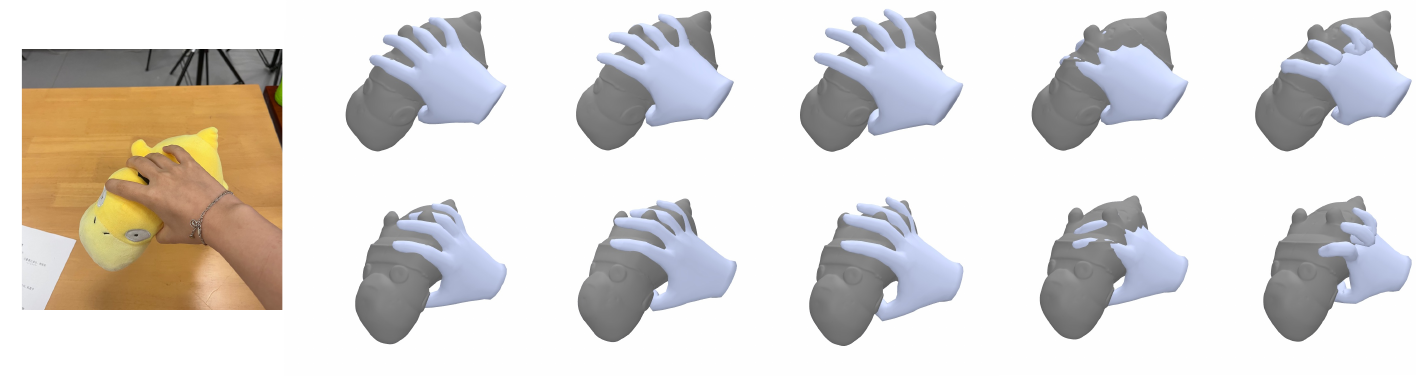}
        \put(6,0){\small Input Image}
        \put(27,0){\small Full Losses}
        \put(41,0){\small w/o $\mathcal{L}_{\text{hand-mask}}$}
        \put(58,0){\small w/o $\mathcal{L}_{\text{contact}}$}
        \put(75,0){\small w/o $\mathcal{L}_{\text{penetr}}$}
        \put(91,0){\small w/o $\mathcal{L}_{\text{reg}}$}

    \end{overpic}
    \caption{ A visualization of the ablation study on loss terms in hand parameter refinement. Each loss term was individually removed from the total loss function, and hand parameter refinement was performed to observe the resulting effects. The top row shows the input viewpoint, while the bottom row provides an alternative viewpoint to more clearly illustrate the differences.}
    \label{fig:loss_ablation}
    \vspace{-1em}
\end{figure*}

\subsection{Comparison Results}
\noindent \textbf{Quantitative Results.}
We evaluate our method's performance in terms of object reconstruction accuracy, grasping quality and hand accuracy. To ensure fair comparison, we employ HaMeR for hand pose detection when evaluating all baseline methods. As shown in \cref{tab:object_recon}, our method achieves the lowest Chamfer distance and highest F-scores. These metrics demonstrate our method's capability to accurately reconstruct objects from single-view images.
In terms of hand-object interaction quality, \cref{tab:object_recon} shows that our method outperforms competing approaches, consistently achieving the lowest simulation displacement and smallest intersection volume across all datasets. These results underscore the effectiveness of our hand-object interaction modeling. 
In~\cref{tab:hand_metric}, we report the reconstruction errors for hand vertices and joints. Since the object geometry recovered by LRM is not entirely accurate, optimizing for physically plausible HOI results with such imperfect object geometry inevitably compromises hand accuracy. Consequently, our final hand results are slightly worse than HaMER's. However, with ground truth object geometry, our proposed HOI optimization can further refine the initial hand results produced by HaMER, demonstrating the effectiveness of our method.

\noindent \textbf{Qualitative Results.}
We evaluate our method through visual comparisons with two baselines, IHOI and MOHO. \cref{fig:gallery_3datasets} showcases representative results on Arctic, OakInk, and DexYCB datasets, demonstrating our method's effectiveness across varied hand-object interactions. Our approach successfully reconstructs hand-object interactions in challenging scenarios with significant occlusions and complex hand poses, highlighting its robustness for single-view hand-object reconstruction in real-world settings. 

The supplementary material contains additional examples from the benchmark datasets that demonstrate consistent performance across diverse scenarios. We also present results on our collected "in-the-wild" images, which showcase our method's robustness in various real-world environments. Furthermore, we provide a qualitative comparison on the selection of large-scale reconstruction models, which shows that integrating our method with Tripo3D~\cite{tripo3d} yields enhanced object reconstruction results and brings notable improvements in hand-object interaction results. This successful integration demonstrates our framework's extensibility and suggests that our approach can effectively leverage future advances in large-scale reconstruction models to achieve even better performance.

\subsection{Ablation Study and Discussions}
\begin{table}[htp]
    \centering
    \resizebox{\linewidth}{!}{
    \begin{tabular}{ccccccc}
        \hline
        & \multicolumn{2}{c}{
        S.D.$\downarrow$
        } & \multicolumn{2}{c}{
        I.V.$\downarrow$
        }
        & \multirow{2}{*}{MPVPE$\downarrow$}  & \multirow{2}{*}{MPJPE$\downarrow$}
        \\
        & Mean & Std & Mean & Std & &  \\
        \hline 
        Full losses & \textbf{3.08} & \textbf{2.92} & \textbf{4.11} & 4.77 &\textbf{1.19} &\textbf{1.22}\\
        w/o $\mathcal{L}_{\text{hand-mask}}$ & 3.21  & 3.69  & 4.28 & \textbf{4.69} &1.73 &1.68\\
        w/o $\mathcal{L}_{\text{penetr}}$ & 2.95  & 3.16  & 9.62 & 10.13 & 1.35 &1.33\\
        w/o $\mathcal{L}_{\text{reg}}$ & 3.24  & 3.73  & 4.47 & 4.82 & 4.57 &4.41\\
        w/o $\mathcal{L}_{\text{contact}}$ & 3.94  & 3.81  & 4.26 & 4.79 & 1.30 &1.27 \\
        \hline
    \end{tabular}
    }
    \vspace{-0.5em}
    \caption{Ablation study of each loss term in hand parameter refinement on OakInk dataset.}
    \vspace{-1.5em}
    \label{tab:ablation-loss}

\end{table}
To analyze each component of our proposed method, we conducted a series of ablation experiments to quantify the impact of individual elements on the overall performance of our method. 

\noindent\textbf{Ablation of HOI Optimization Stages.}
To analyze the three stages of HOI optimization scheme, we  quantitatively and qualitatively evaluate the intermediate results on datasets of Arctic, OakInk and DexYCB. 
As illustrated in Fig.~\ref{fig:hand_optim_stage}, each stage of our optimization pipeline yields incremental improvements over its predecessor.
\textit{1) Camera System Setup:}
Before optimization, imprecise camera parameters lead to misaligned object and hand positions, with grasp contacts deviating from the image. The first stage aligns the rendered object with the input image, however it leaves hand inconsistent with the image. 
\textit{2) HOI Contact Alignment:}  The second stage optimizes hand to achieve both precise image alignment and appropriate contacts, but introduces significant hand-object intersections. 
\textit{3) Hand Parameter Refinement:} The final stage ensures physical plausibility and stability of the grasping by eliminating penetrations while preserving fingertip contacts.
The quantitative results of~\cref{tab:ablation-optim-comp} further validates the conclusion that the whole pipeline yields the best results.

\noindent\textbf{Ablation of Loss Terms in Hand Parameter Refinement.}
To further examine Hand Parameter Refinement, we conducted an ablation study on the loss terms, with results shown in ~\cref{fig:loss_ablation} and~\cref{tab:ablation-loss}. Both qualitative and quantitative evaluations offer insights into the significance of each loss term. Although $\mathcal{L}_{\text{hand-mask}}$ was optimized in a prior step, its removal still increases simulation displacement and introduces deviations of hand shape from the input image. Additionally, removing $\mathcal{L}_{\text{penetr}}$ results in more intersections, highlighting its essential role in maintaining physical realism. Without $\mathcal{L}_{\text{reg}}$, we observe twisted fingers, higher simulation displacement, and larger intersection volumes, which reduce grasp fidelity. Lastly, omitting $\mathcal{L}_{\text{contact}}$ positions the fingertips farther from the object surface and increases simulation displacement, underscoring its critical role in preserving grasp stability. These findings confirm that each loss term is crucial for achieving physically plausible and stable hand-object interactions.
\section{Conclusion}
In this paper, we explore the use of large-scale models to reconstruct HOI from single-view RGB images. We start by initial reconstruction of hand and object and propose a prior-guided optimization scheme that ensures physical plausibility while preserving the visual fidelity of the input image. Extensive experiments on both public datasets and our own collected data validate the effectiveness of our approach, demonstrating superior generalization capabilities and underscoring the potential of large-scale models in HOI reconstruction.
As our method relies heavily on the adopted large models, it is susceptible to unsatisfactory results when these models fail. In future work, we aim to enhance efficiency and robustness by exploring how to improve large models to specifically address the unique requirements of hand-object interaction tasks.

\section{Acknowledgements}
This work was supported by NSFC (No.62206173), Shanghai Frontiers Science Center of Human-centered Artificial Intelligence (ShangHAI), MoE Key Laboratory of Intelligent Perception and Human-Machine Collaboration (KLIP-HuMaCo). The authors thank Qi Jiang and Yaxun Yang (ShanghaiTech University) for their helpful discussions on this project.
{
    \small
    \bibliographystyle{ieeenat_fullname}
    \bibliography{main}
}

\clearpage
\setcounter{page}{1}
\maketitlesupplementary

This supplementary material provides additional details on our method and results that complement the main paper. \cref{sec:init-recon} details the segmentation and reconstruction of the object. \cref{sec:hoi-optim} elaborates on the HOI contact alignment stage of the HOI optimization process. Finally, \cref{sec:exp} presents further experiments and analyses, demonstrating the robustness and versatility of our method.
\section{Initial Reconstruction of Hand and Object}
\label{sec:init-recon}
\subsection{Hand-Object Interaction Reasoning}
\begin{figure}[htp]
    \centering
    \includegraphics[width=\linewidth]{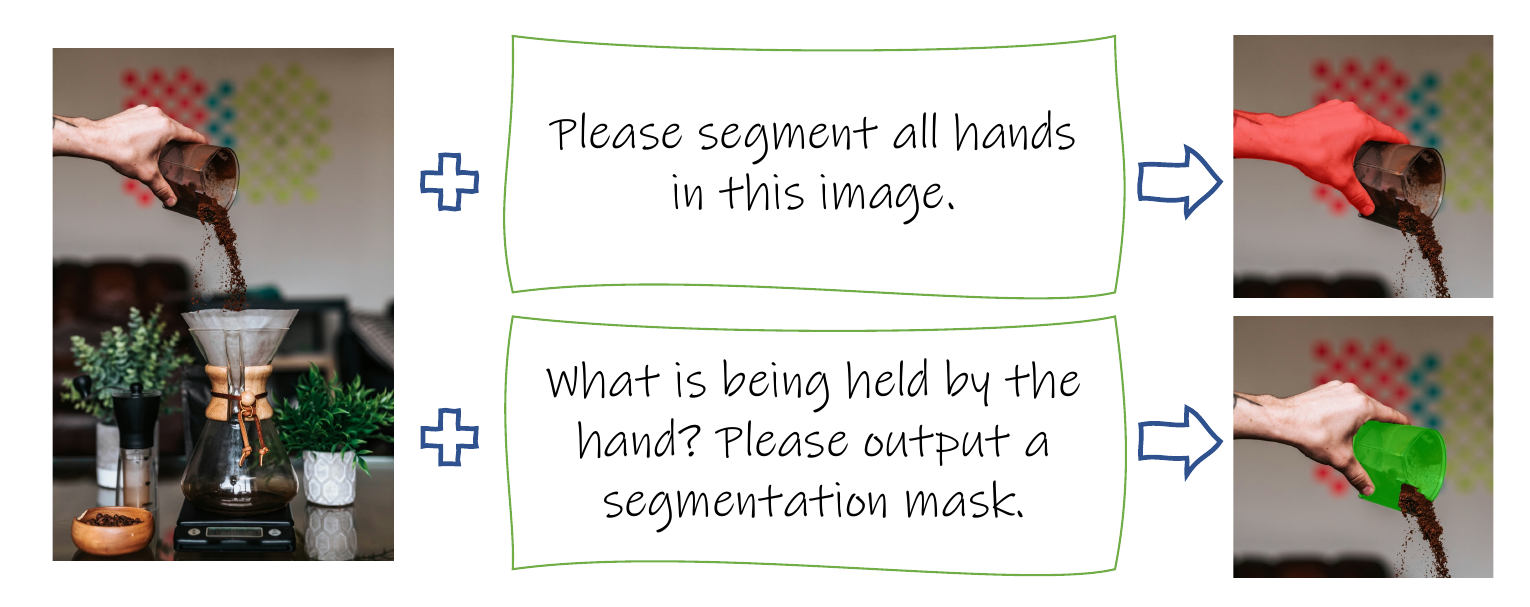}
    \caption{Given an input image, we use predefined prompt to reason the segmentation of hand and object.}
    \label{fig:vlm-result}
\end{figure}

\begin{figure}
    \centering
    \includegraphics[width=\linewidth]{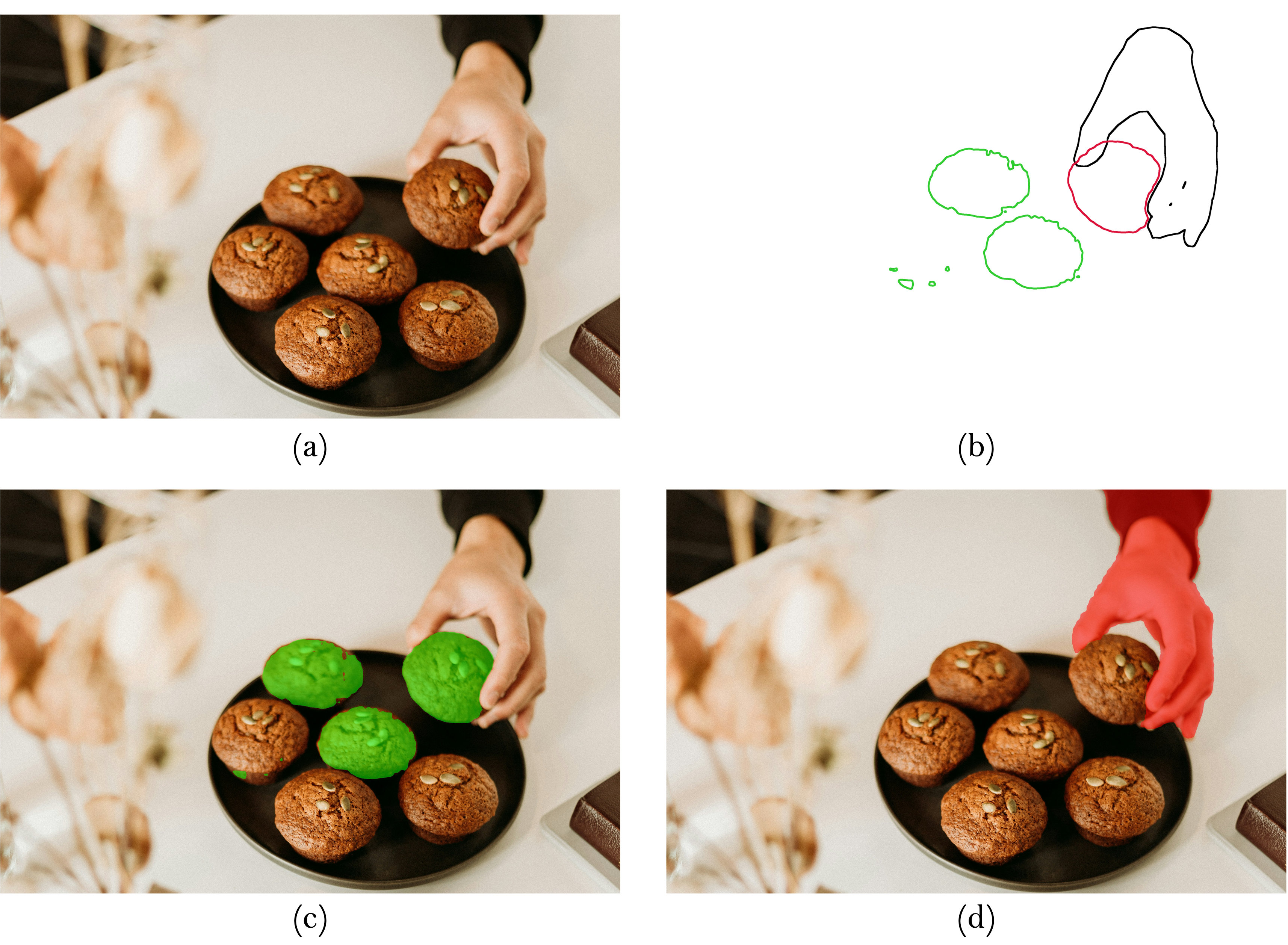}
    \caption{The figure illustrates the segmentation and contour extraction for hand-object interaction analysis. Image (a) is the input image. Image (b) displays the corresponding contours extracted from the object and hand masks. Black contours represent the hand, while red contours highlight the target object parts crucial for HOI understanding. Green contours indicate redundant masks identified for removal, as they do not contribute to the hand-object interaction being analyzed. Image (c) and (d) depict the segmented object and hand masks.  }
    \label{fig:seg_contour}
\end{figure}
Before reconstructing the Hand-Object interaction, we first need to identify the region of interest, specifically, the area in the input image where the object is in interaction with the hand. This is a challenging task, as many in-the-wild images contain multiple objects, but only one is being actively interacted with.

\paragraph{Reasoning with Vision-Language Model.} Inspired by the recent success of vision-language models in image understanding, we employ LISA, a context-aware segmentation model, to analyze and segment hand-object interactions. As illustrated in~\cref{fig:vlm-result}, given a single input image, we prompt the LISA model with two queries to obtain segmentation masks for the hand and the object: 1) \textit{"Please segment all hands in this image."}; 2) \textit{"What is being held by the hand? Please provide its segmentation mask."} LISA's visual-language capabilities enable precise segmentation masks for both the hand and its interacting object. 

\paragraph{Contour-guided Filtering.}
Although the LISA model can successfully reason about and segment the hand and the object it interacts with in most cases, we observed there still exist imperfections in the segmentation masks that hinder further processing. 
As shown in~\cref{fig:seg_contour}, LISA incorrectly segments redundant masks of objects that are not interacted with hands. This error may arise because the cookies share the same language description and similar visual appearance.

To address the issue mentioned above, we propose a contour-guided filtering strategy. Specifically, we first extract the contours of the hand and all segmented objects. If an object is being interacted with, its contour should be adjacent to the hand's. Based on this assumption, we discard objects whose contours are not neighboring the hand's. This approach enables us to accurately obtain the segmentation mask for the objects that hands interacts with.

\subsection{Object Reconstruction}
\label{suppl:obj-recon}
\begin{figure}[htp]
    \centering
    \includegraphics[width=\linewidth]{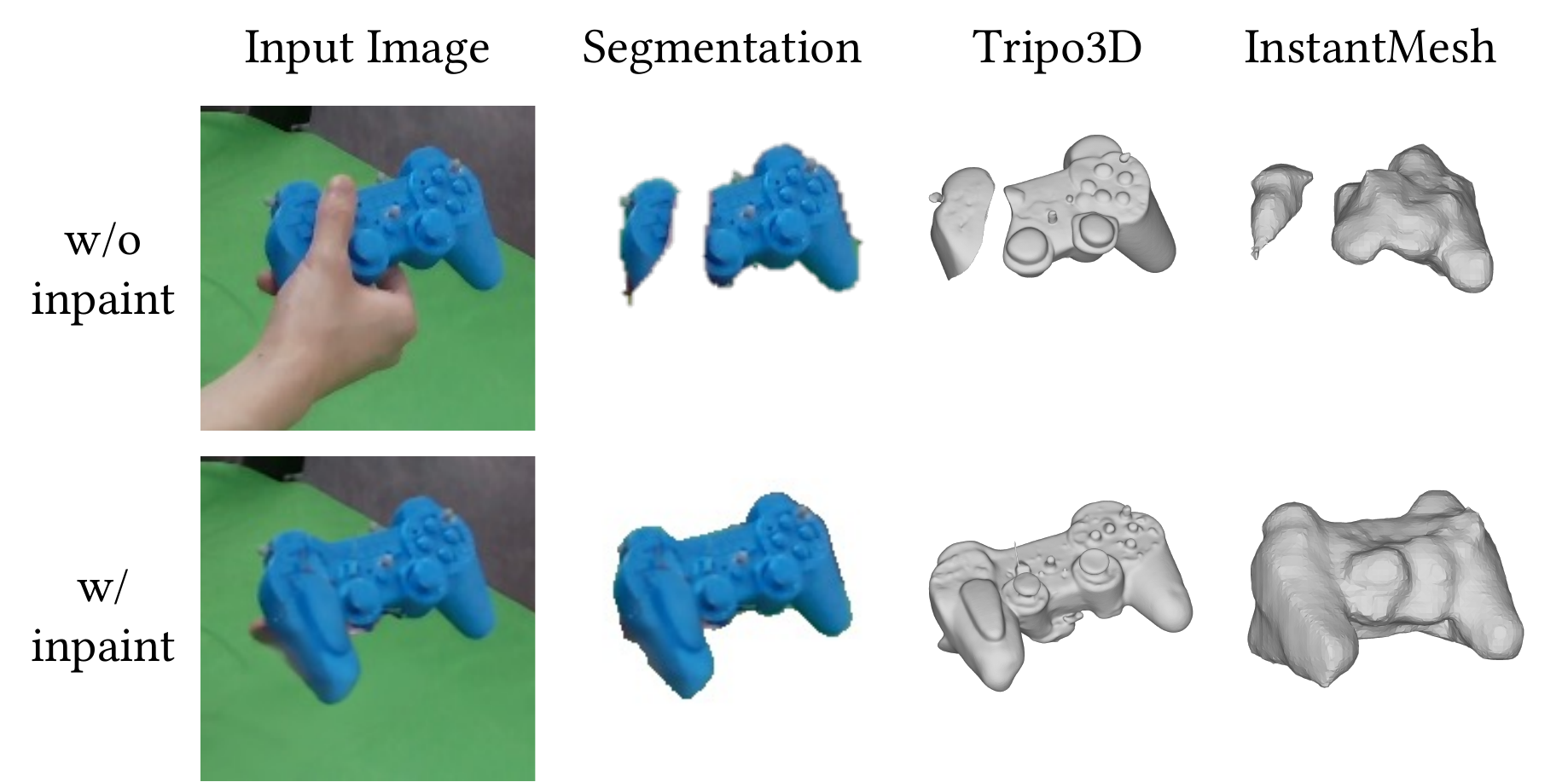}
    \caption{We conducted a comparative analysis of reconstruction results between original images and those subjected to inpainting. The top row displays results from the original image, while the bottom row presents results obtained from images after applying the inpainting process.}
    \label{fig:inpaint-ablation}
\end{figure}

\begin{figure}
    \centering
    \includegraphics[width=\linewidth]{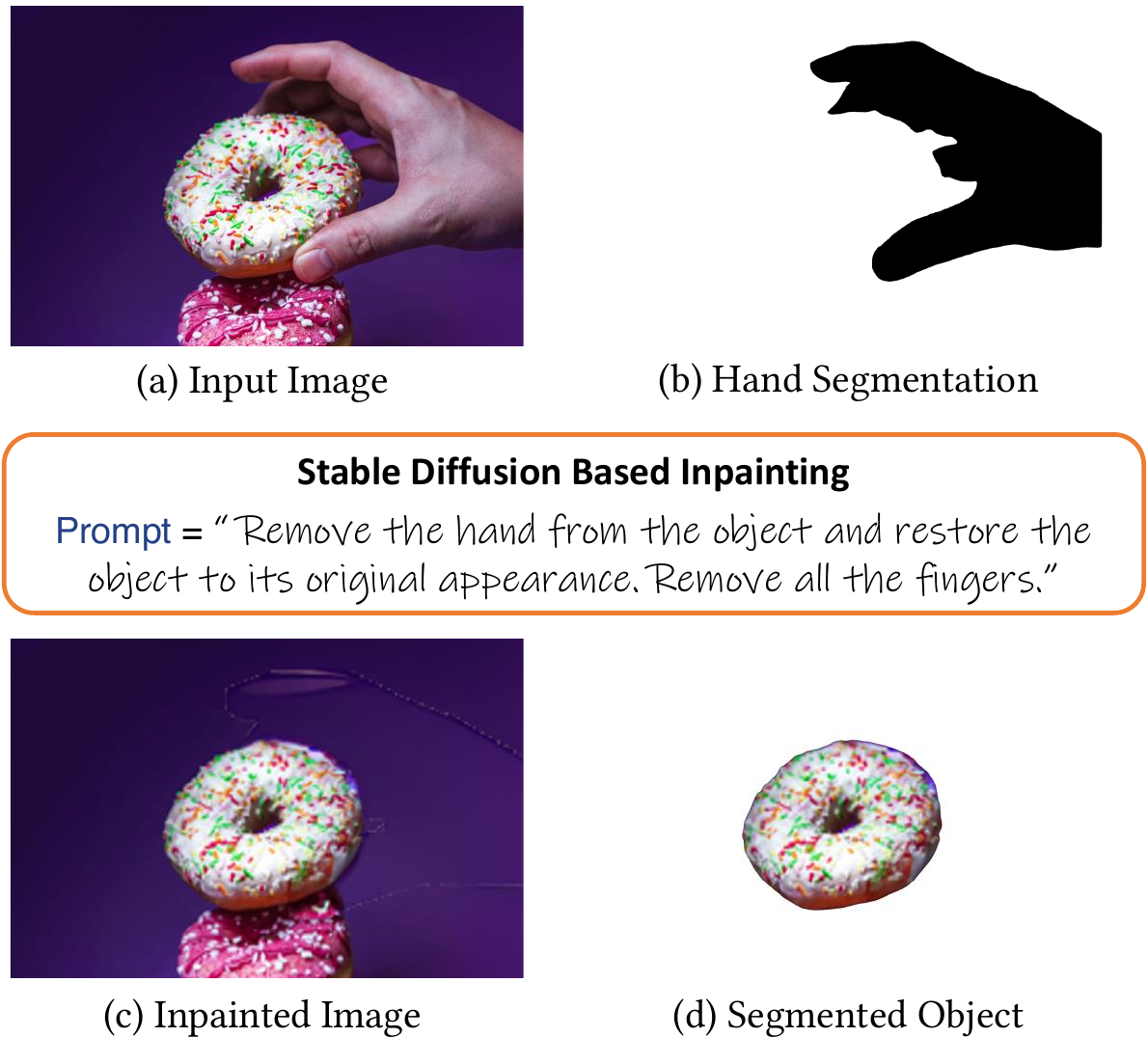}
    \caption{Illustration of the inpainting process. Given an input image containing a hand and a corresponding hand mask, a text-guided diffusion model effectively removes the hand from the image and inpaints the masked region.}
    \label{fig:inpainting}
\end{figure}
Here we present details on how to reconstruct the object from input image. First we remove occlusions from the image, then re-segment the complete object image, and finally generate the corresponding object mesh using this object image.
\paragraph{Object occlusion removal via image inpainting.}
Since objects interacting with hands are often partially occluded in the image, directly using the original input to reconstruct the object's 3D geometry can result in distorted and incomplete shapes. To obtain a more accurate 3D geometry, we first use a diffusion model~\cite{ye2023affordance, nichol2021glide} to recover the complete appearance of the object in 2D image.

As illustrated in~\cref{fig:inpainting}, we employ a stable diffusion model for object inpainting, using the input image and hand mask alongside a tailored text prompt. The hand mask identifies regions requiring inpainting, while the text prompt guides the reconstruction of the object's original appearance. Thanks to the robust generalization capabilities of stable diffusion, this inpainting approach successfully synthesizes the occluded object regions across diverse scenarios, producing photorealistic results.

\paragraph{Re-segment from Inpainted Image.}
With the inpainted image, we utilize a large reconstruction model, InstantMesh, to reconstruct the object's geometry. Since InstantMesh requires a background-free input, we must first obtain the segmentation mask of the inpainted object. To generate this mask, we use the occluded object mask as an indicator.
As shown in~\cref{fig:inpaint-ablation}, the occluded mask typically consists of multiple sub-masks due to the hand separating the object. We randomly sample points within each sub-mask and compute a bounding box that loosely covers the occluded mask. These sampled points and the bounding box serve as prompts for the SAM model, which extract the object from the inpainted image. Finally, InstantMesh takes the completed object as input and reconstructs its geometry.

\paragraph{Watertight Post-processing.}
In hand-object interactions, mutual occlusions naturally occur and are intrinsically linked to contact relationships. The reconstructed meshes from the LRM are sometimes non-watertight, which can hinder robust and accurate hand-pose optimization. To address this, we convert the non-watertight meshes into watertight ones when needed.
For a non-watertight mesh, we first render depth maps from multiple viewpoints that cover the entire object. These depth maps are then fused into a unified point cloud, which helps eliminate isolated and occluded parts. Next, we apply the Poisson reconstruction method to generate a mesh from the point cloud. Finally, a hole-filling algorithm~\cite{attene2010lightweight} is used to ensure the mesh meets the watertight requirement.

\section{Hand-Object Interaction Optimization}
\label{sec:hoi-optim}
\paragraph{HOI Contact Alignment.}
We identify potential contact regions by analyzing two types of hand-object overlaps in the input image. For front-side contacts, where the object is occluded by the hand, we compute the contact mask $M_{\text{front}} = \hat{M}_o \setminus M_o$ as the difference between the inpainted object mask $\hat{M}_o$ and the original object mask $M_o$. For back-side contacts, where the hand is occluded by the object, we derive the contact mask $M_{\text{back}} = \hat{M}_h \setminus M_h$ as the difference between complete hand mask $\hat{M}_h$ and the segmented hand mask $M_h$. The complete hand mask is obtained by rendering on the pose and camera parameters estimated by HaMeR. 

From the contact masks $M_{\text{front}}$ and $M_{\text{back}}$, we recover 3D contact points via ray-casting to hand and object geometries seperately. 
As shown in the Fig. 4 in the main paper, we can emit a ray from each pixel on contact masks to hit the reconstructed object and hand. Through the application of rasterization and depth peeling techniques, we extract multiple depth values from different layers of the 3D models. In our implementation, we utilize four depth layers, which we have empirically found to be sufficient for all test cases in our experiments.

For ray-object intersections, we select the minimum depth values within $M_{\text{front}}$ and maximum depth values within $M_{\text{back}}$, corresponding to the nearest and farthest points from the camera respectively.

Regarding the ray-hand intersection, it is important to note that the functional area for grasping is limited to the palmar surface. The dorsal side of the hand, comprising the back of the hand and fingers, is not involved in object manipulation. We manually select and label faces corresponding to the palmar and dorsal regions on the MANO template model as a preprocessing step. This anatomical annotation serves as prior knowledge, allowing us to efficiently exclude 3D points located on the dorsal side. Therefore, valid hand contact points are determined for each pixel in $M_{\text{front}}$ and $M_{\text{back}}$ by filtering ray intersections based on face indices to retain only palmar-side points, then selecting the nearest and farthest intersections based on depth value.

Once all potential contact points on both the hand and the object are identified, we apply the Iterative Closest Point (ICP) method to compute the optimal hand translation, aligning the contact points and providing a rough estimation of the hand's pose.

\section{Experiments}
\label{sec:exp}
\subsection{Running Time Analysis}
We measured the processing time for 65 images at each stage and calculated the average runtime. The initial reconstruction stage required an average of 58.97 seconds, with the breakdown as follows: HOI Reasoning (11.68 seconds), Hand Reconstruction (5.54 seconds), Object Inpainting (15.55 seconds), and Object Reconstruction (26.20 seconds). For the optimization stage, the average runtime was 57.03 seconds, consisting of Camera Setup (4.41 seconds), Contact Alignment (30.51 seconds), and Hand Refinement (22.11 seconds).

\subsection{Ablation on Large Models }
\begin{figure}[h]
    \centering
    \scriptsize
    \begin{overpic}[width=\linewidth]{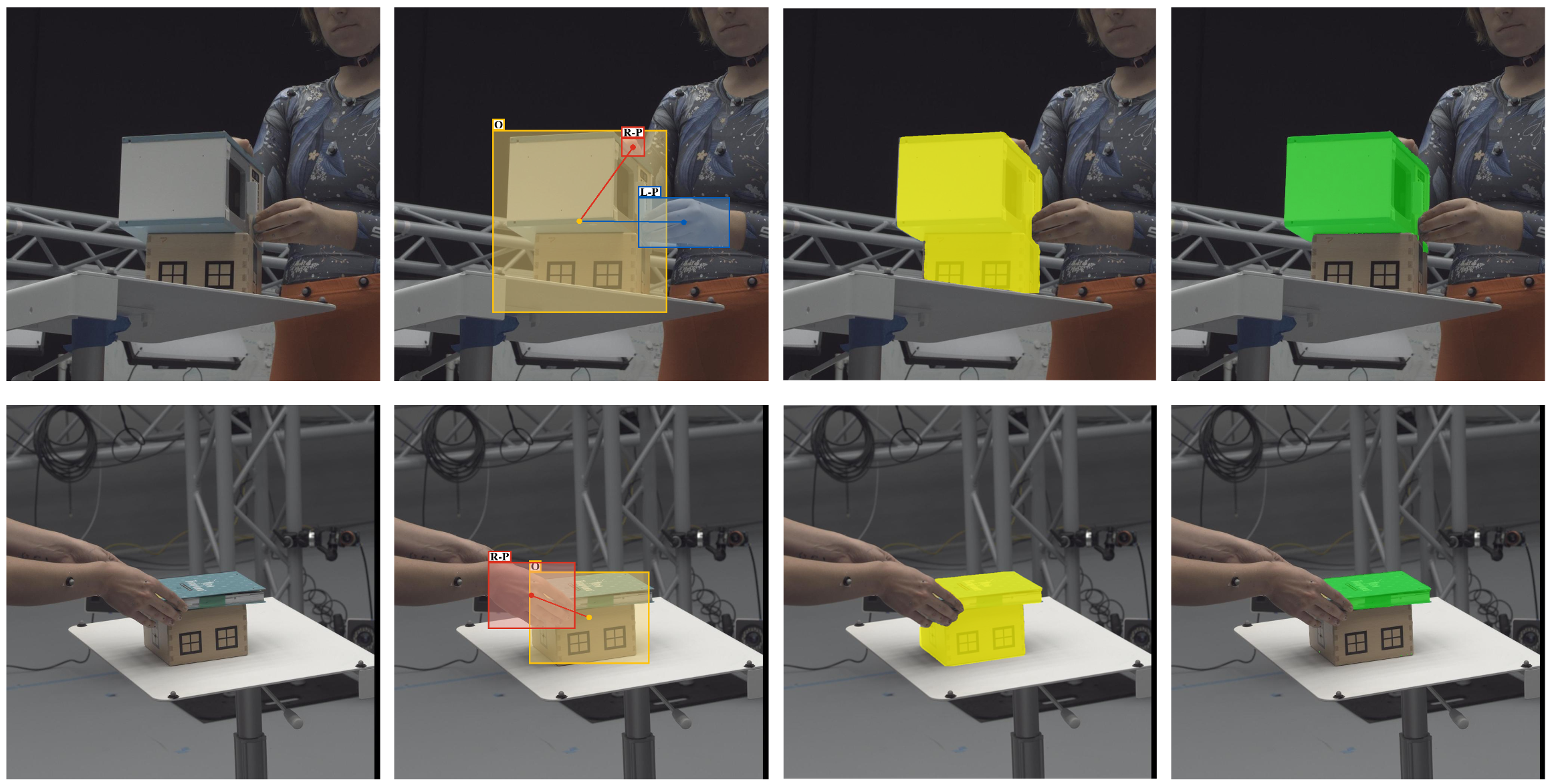} 
        \put(8, -3){Input}
        \put(34, -3){HODet}
        \put(55, -3){HODet+SAM}
        \put(85, -3){LISA}      
    \end{overpic}
    \caption{Visualization of the ablation on HOI reasoning large model. }
    \label{fig:hoi-reasoning}
\end{figure}

\paragraph{The Selection of HOI Reasoning Models.} 
In addition to the vision-language model, we examine the Hand-Object detector (HODet)\cite{shan2020understanding} for segmenting hands and objects in input images. To compare HOI reasoning performance, we evaluate LISA against HODet on the Arctic dataset. As HODet predicts only object bounding boxes, we employ SAM to generate object segmentations based on these predictions. LISA outperforms HODet, achieving an average IoU of 0.74 compared to 0.61. A visual comparison in~\autoref{fig:hoi-reasoning} shows that HODet frequently misidentifies objects and detects extraneous elements.

\paragraph{The Selection of Large Reconstruction Models.} While our pipeline incorporates the open-source model InstantMesh for object reconstruction, it could significantly benefit from a more advanced model. For comparison, we employ the state-of-the-art commercial model Tripo3D~\cite{tripo3d}. \cref{fig:gallery} displays the reconstructed meshes produced by both approaches on a range of challenging in-the-wild images. This comparison highlights the potential of our approach to combine the strengths of multiple large-scale models to achieve highly accurate object reconstruction across diverse scenarios.

To assess the impact of reconstruction quality, we evaluate the optimization results using InstantMesh reconstruction, Tripo3D reconstruction, and ground truth (GT) meshes on Oakink dataset. As shown in the table below, lower-quality reconstructions (indicated by higher Chamfer Distance (C.D.) values) result in poorer HOI performance.

\begin{table}[htpb]
    \centering
    \resizebox{\linewidth}{!}{
\begin{tabular}{cccc|cc|cc}
\hline 
& \multicolumn{3}{c|}{ Object quality } & \multicolumn{2}{c|}{ HOI results } & \multicolumn{2}{c}{ final hand results }  \\
& $\mathrm{F} 5 \uparrow$ & $\mathrm{F} 10 \uparrow$ & $C.D.\downarrow$ &S.D.$\downarrow$ & I.V.$\downarrow$ & MPVPE$\downarrow$ & MPJPE$\downarrow$ \\
\hline 
GT  &\textbf{1.00}   &\textbf{1.00}  &\textbf{0.00} &\textbf{1.92} &\textbf{2.44} & \textbf{0.86} & \textbf{0.77} \\
Tripo3D &\underline{0.280} &\underline{0.503}  &\underline{0.842} &\underline{2.76} & \underline{4.05} &\underline{1.15} & \underline{1.21} \\
InstantMesh &0.247 & 0.445 & 1.035 & 3.08 &4.11  & 1.19& 1.22 \\

\hline
\end{tabular}
}
\caption{The impact of object reconstruction quality on HOI optimization performance.}
\label{tab:robustness}
\end{table}

\subsection{Qualitative Comparison Results}
\begin{figure*}[htp]
    \centering
\includegraphics[width=\linewidth]{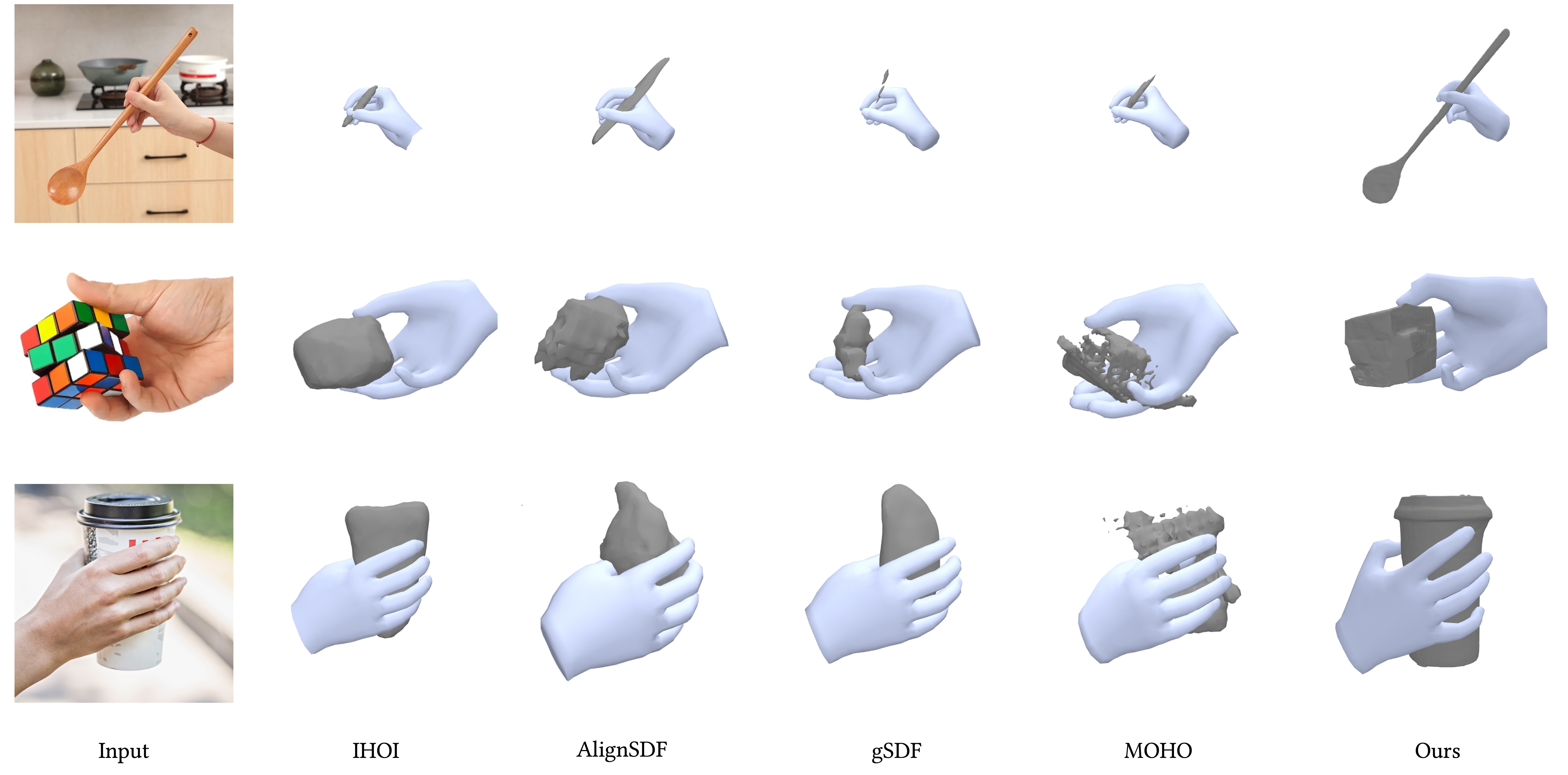}
    \caption{Qualitative Comparison of IHOI, AlignSDF, gSDF, MOHO, and Our Method on in-the-wild Images}
    \label{fig:compare-wild}
\end{figure*}

\paragraph{Comparative results on In-th-wild Images} As depicted in~\autoref{fig:compare-wild}, we showcase qualitative comparisons with IHOI, AlignSDF, gSDF, MOHO, and our method on in-the-wild images. Our approach excels in accurately reconstructing intricate object geometries and details.
\paragraph{Additional comparative results on public datasets.} Here we provide additional comparative results on public datasets. \cref{fig:gallery-oakink} demonstrates comparisons with IHOI and MOHO on the OakInk dataset, while \cref{fig:gallery-arctic} and \cref{fig:gallery-dexycb} show our method's performance on the Arctic and DexYCB datasets, respectively. These additional examples demonstrate our method's performance across diverse scenarios.

\subsection{Failure Cases Analysis}
\begin{figure}[h]
    \centering
    \includegraphics[width=\linewidth]{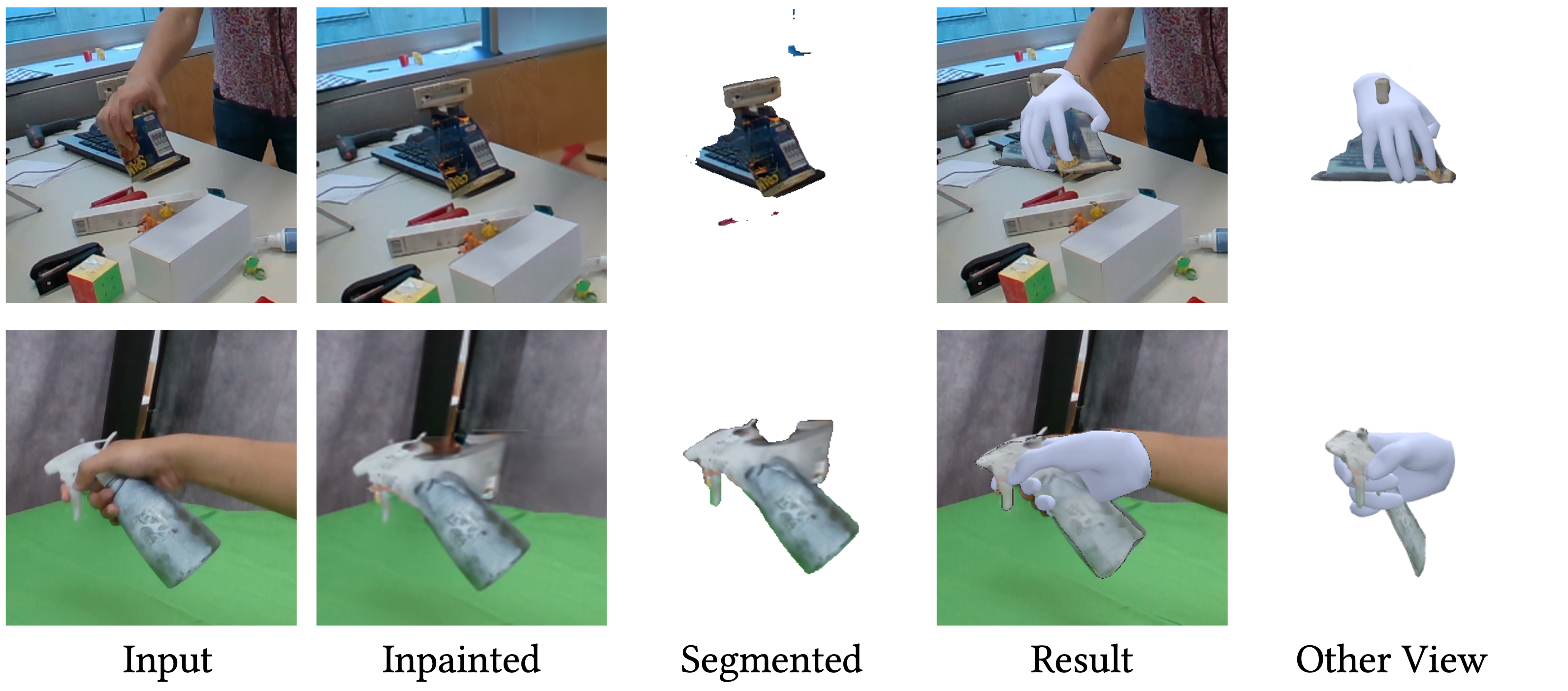}
    \caption{Examples of failure Cases in the OakInk dataset. Failures primarily result from artifacts introduced by the inpainting model.}
    \label{fig:failure-cases}
\end{figure}

Most failures originate from the initial reconstruction stage rather than the optimization stage. As shown in~\autoref{fig:failure-cases}, the inpainting model occasionally introduces artifacts, causing the object being held to blend with the background and making segmentation challenging. Consequently, the input images for the large reconstruction model become incoherent and deviate significantly from real-world objects. We believe that incorporating more advanced inpainting models and leveraging hand contours in the segmentation process are promising directions for future exploration.

\begin{figure}[h]
    \centering
\includegraphics[width=\linewidth]{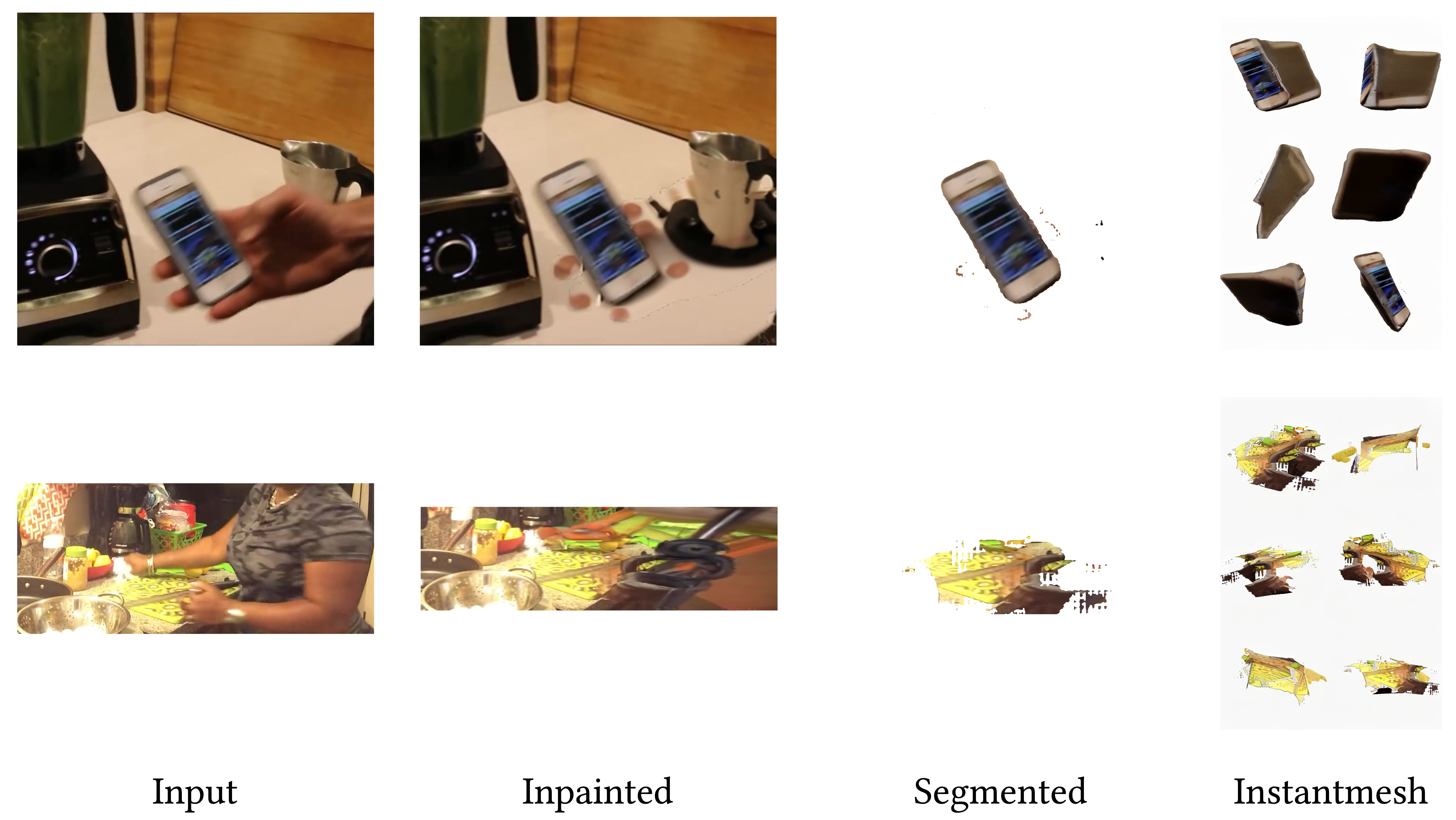}
    \caption{Examples of failure cases on the MOW dataset. The last column is the synthesized 6 views from InstantMesh, it shows that motion blur and ambiguous grasping present significant challenges for object reconstruction.}
    \label{fig:mow-full}
\end{figure}

\begin{figure*}[htpb]
    \centering
    \includegraphics[width=\linewidth]{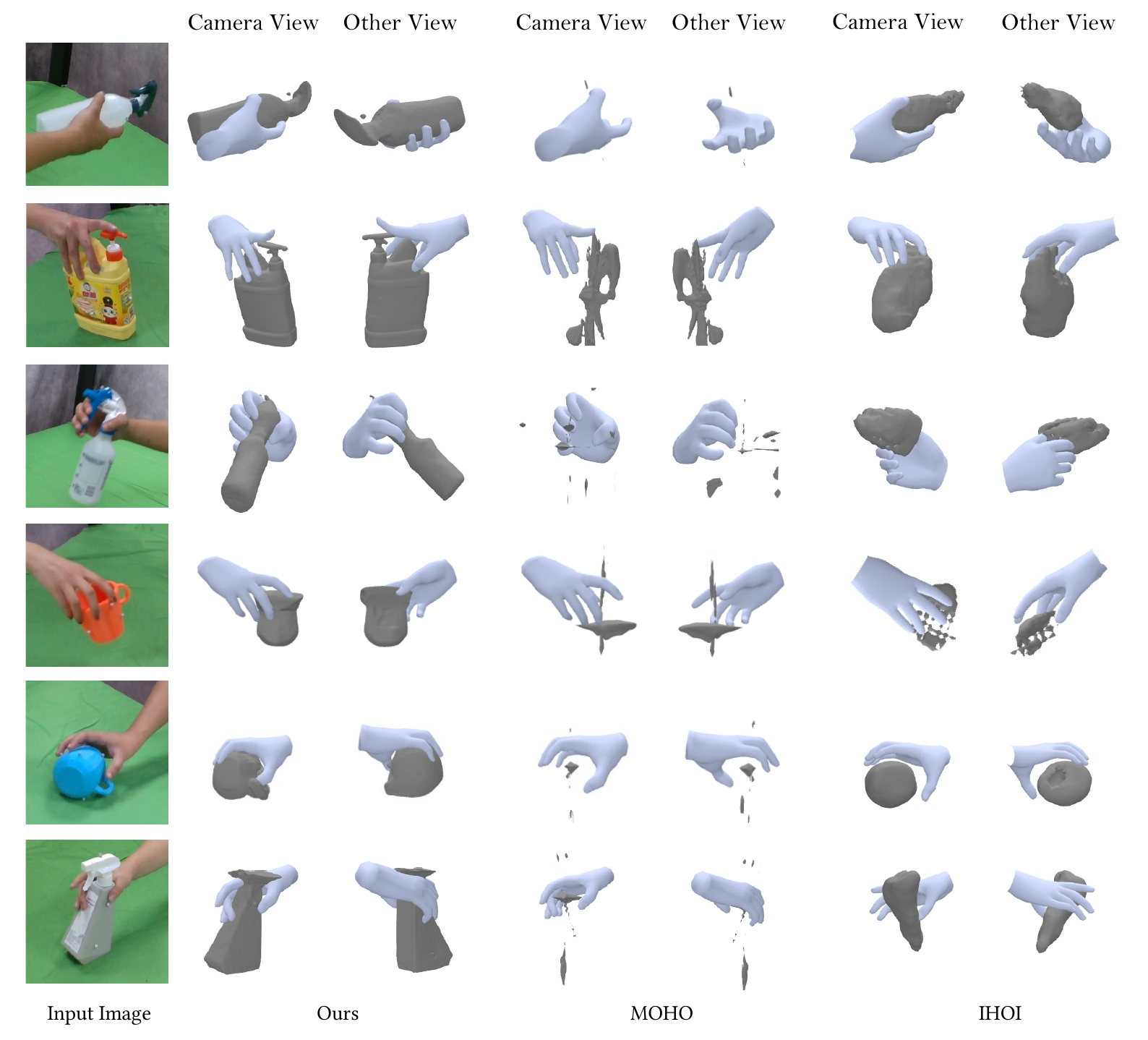}
    \caption{This gallery showcases the outcomes of our hand-object reconstruction on the dataset OakInk. The first column is the input image, we present the camera view and another view to display the reconstructed HOI meshes.}
    \label{fig:gallery-oakink}
\end{figure*}

\begin{figure*}[htbp]
    \centering
    \begin{minipage}{\textwidth}
        \centering
        \includegraphics[width=\linewidth]{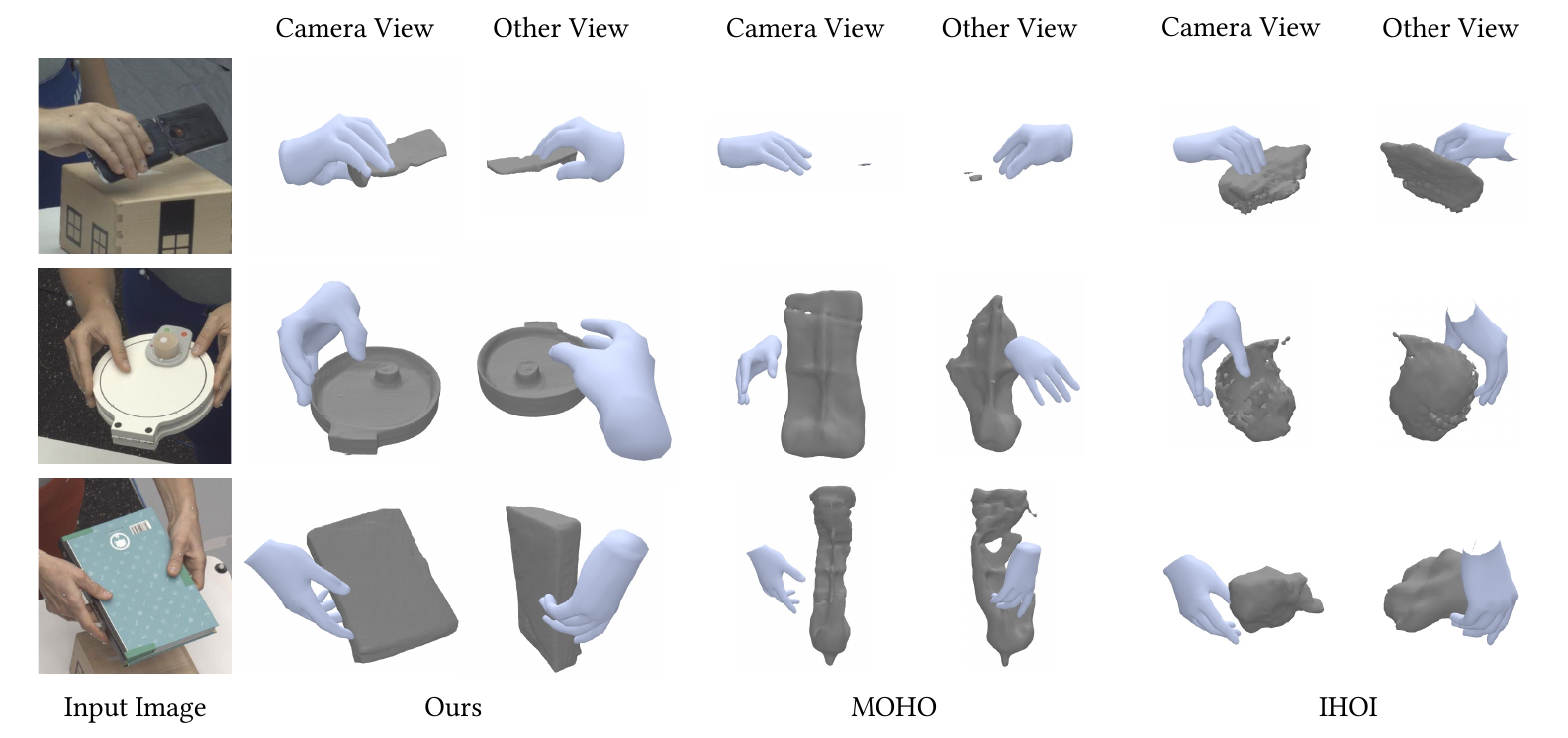}
        \caption{This gallery showcases the outcomes of our hand-object reconstruction on the dataset Arctic. }
        \label{fig:gallery-arctic}
    \end{minipage}
    \hfill
    \hfill
    \begin{minipage}{\textwidth}
        \centering
        \includegraphics[width=\linewidth]{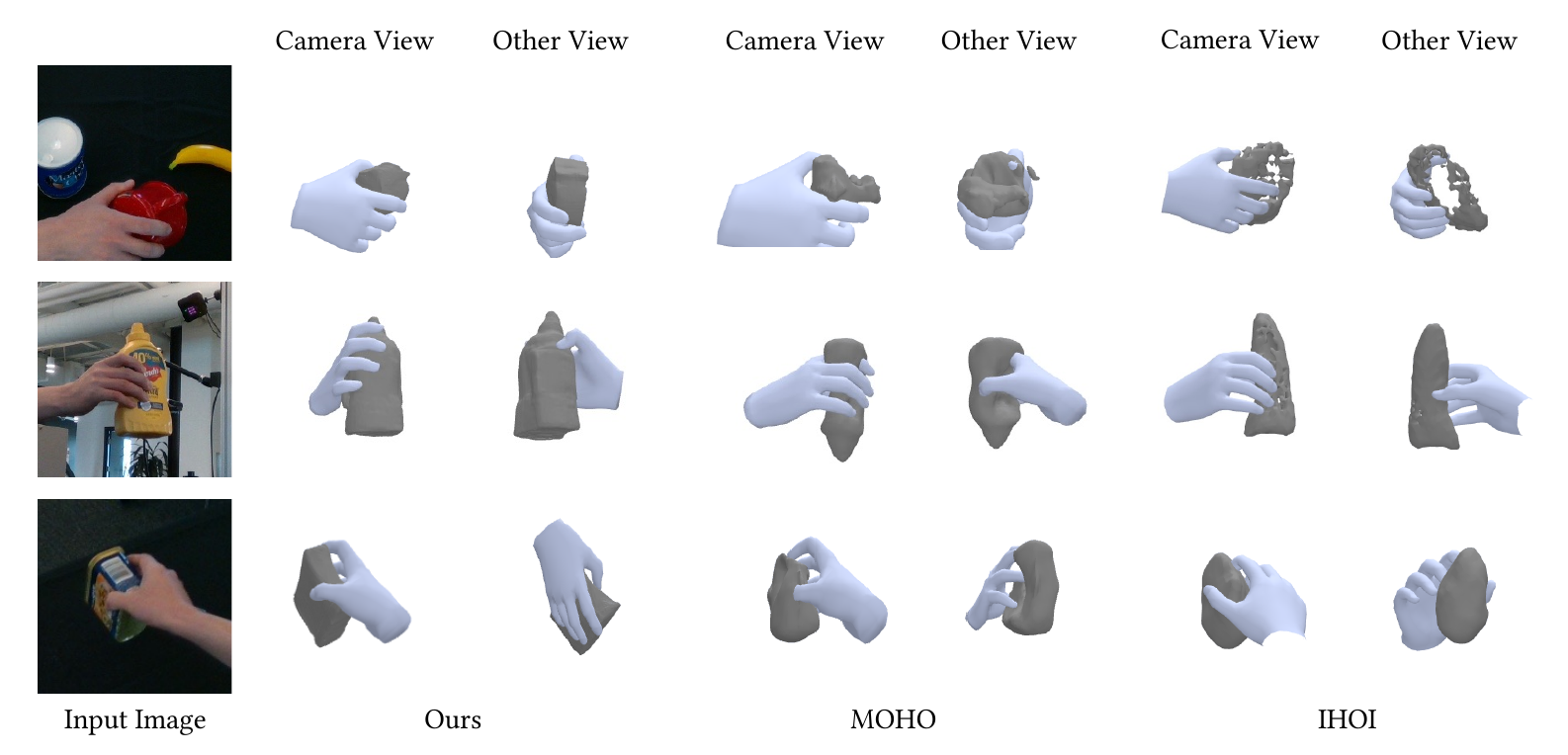}
        \caption{This gallery showcases the outcomes of our hand-object reconstruction on the dataset DexYCB.}
        \label{fig:gallery-dexycb}
    \end{minipage}
\end{figure*}

\begin{figure*}[htpb]
    \centering
    \includegraphics[width=\linewidth]{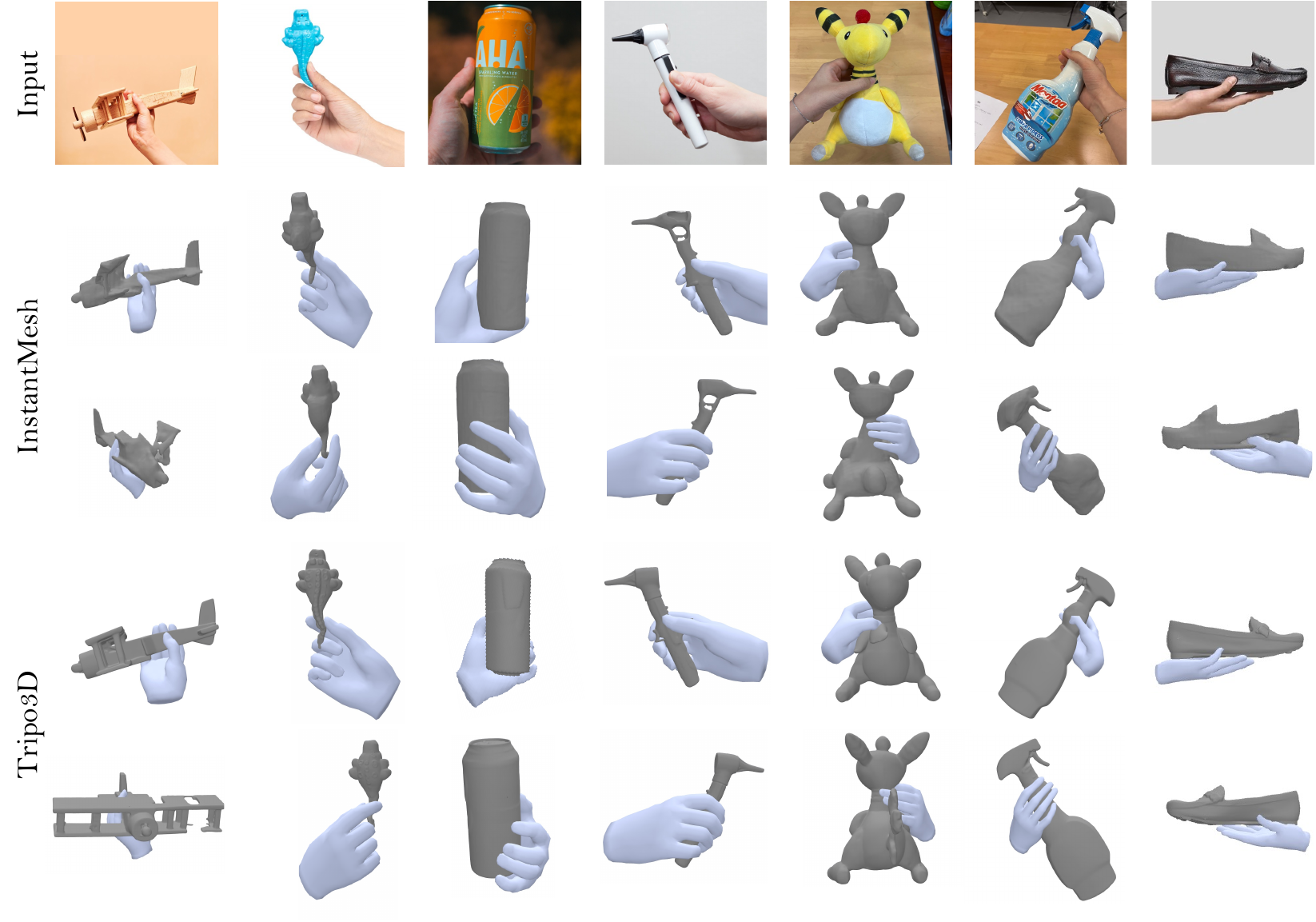}
    \caption{This gallery showcases the outcomes of our hand-object reconstruction results on in-the-wild images, we test the reconstruction result on two LRM, instantmesh and tripo3d.}
    \label{fig:gallery}
\end{figure*}

We also evaluated our pipeline on the MOW dataset. The images in this dataset are of relatively low quality, frequently displaying ambiguous grasping poses and motion blur. These factors present significant challenges for both inpainting and object reconstruction. \autoref{fig:mow-full} illustrates typical examples of failure cases observed on the MOW dataset.

\end{document}